\documentclass[11pt]{article}

\usepackage[accent=orange,notitlelogo]{pangram}
\renewcommand{\footrulewidth}{0pt}
\usepackage{amsmath,amsfonts,bm}

\def\eqref#1{equation~\ref{#1}}

\def\1{\bm{1}}

\DeclareMathAlphabet{\mathsfit}{\encodingdefault}{\sfdefault}{m}{sl}
\SetMathAlphabet{\mathsfit}{bold}{\encodingdefault}{\sfdefault}{bx}{n}

\usepackage{amsmath}
\usepackage{amssymb}
\usepackage{graphicx}
\usepackage{colortbl}
\usepackage{array}
\usepackage{booktabs}
\usepackage{wrapfig}
\usepackage{xspace}
\usepackage[normalem]{ulem}
\newif\iffinal
\iffinal
\newcommand{\todo}[1]{\unskip}
\newcommand{\mohit}[1]{\unskip}
\newcommand{\jenna}[1]{\unskip}
\newcommand{\john}[1]{\unskip}
\newcommand{\katherine}[1]{\unskip}
\newcommand{\bradley}[1]{\unskip}
\newcommand{\ben}[1]{\unskip}

\renewcommand{\sout}[1]{}
\else
\definecolor{lightred}{HTML}{e99090}
\newcommand{\todo}[1]{{\color{lightred}\{\textit{#1}\}$_{TODO}$}}
\newcommand{\mohit}[1]{{\color{cyan}\{\textit{#1}\}$_{mohit}$}}
\newcommand{\jenna}[1]{{\color{teal}\{\textit{#1}\}$_{jenna}$}}
\newcommand{\john}[1]{{\color{violet}\{\textit{#1}\}$_{john}$}}
\newcommand{\katherine}[1]{{\color{purple}\{\textit{#1}\}$_{katherine}$}}
\newcommand{\bradley}[1]{{\color{green}\{\textit{#1}\}$_{bradley}$}}
\newcommand{\ben}[1]{{\color{orange}\{\textit{#1}\}$_{ben}$}}
\definecolor{darkorange}{HTML}{CC6600}

\fi
\newcommand{\codelink}{ at \url{https://github.com/pangramlabs/WildAI}}
\newcommand{\figdir}{figures_pangram}

\newcommand{\lawresultstablefile}{tables_pangram/law_results_table_float}
\makeatletter
\titleformat{\paragraph}[runin]{\sffamily\bfseries\normalsize\color{pangram@heading}}{}{0em}{}
\makeatother

\definecolor{wildaiorange}{HTML}{15502E}
\definecolor{wildaipink}{HTML}{FF6106}
\newcommand{\wildai}{{\color{wildaiorange}\textsc{Wild}}{\color{wildaipink}\textsc{AI}}\xspace}

\title{How Much Is an AI Token Worth? Scaling Laws for Wild AI-Generated Web Text}

\DeclareRobustCommand{\pangramaffilmark}{%
  \kern0.08em
  \raisebox{0.45ex}{\includegraphics[height=0.54em]{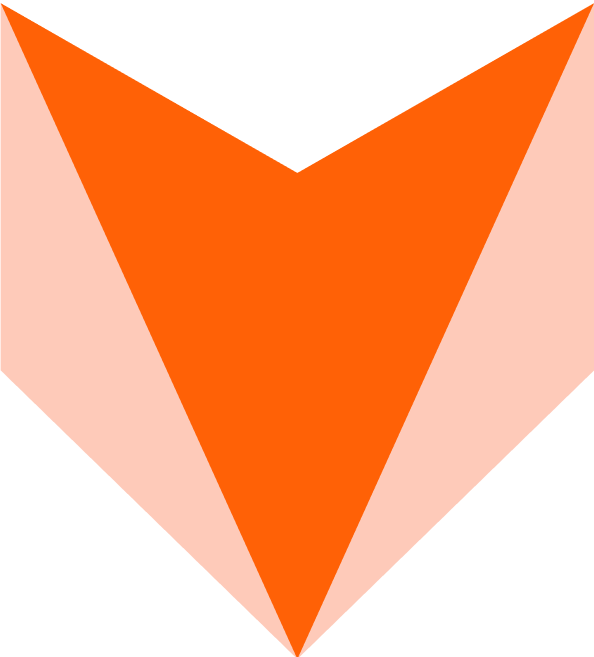}}}
\DeclareRobustCommand{\umdaffilmark}{%
  \kern0.08em
  \raisebox{0.18ex}{\includegraphics[height=0.82em]{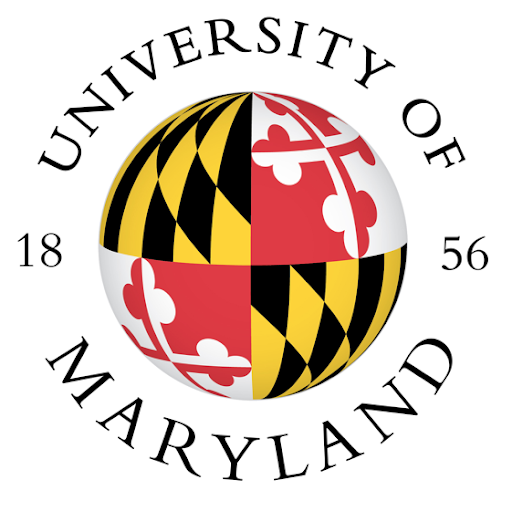}}}

\author{%
  \begin{tabular}{@{}l@{}}
    {\normalfont\Authfont Jenna Russell\umdaffilmark\textsuperscript{*}\enspace
      Benjamin Glickenhaus\pangramaffilmark\enspace
      Katherine Thai\pangramaffilmark}\\%
    {\normalfont\Authfont John Wieting\enspace
      Mohit Iyyer\umdaffilmark\enspace
      Max Spero\pangramaffilmark\enspace
      Bradley Emi\pangramaffilmark}\\%
    {\normalfont\Affilfont
      \umdaffilmark\,University of Maryland\quad
      \pangramaffilmark\,Pangram Labs}\\%
    {\normalfont\sffamily\small
      Correspondence: \href{mailto:jennarus@umd.edu}{jennarus@umd.edu}}
  \end{tabular}%
}

\pangramlogo{pangram-icon.png}
\makeatletter
\renewenvironment{abstract}{%
  \begin{tcolorbox}[
    enhanced, breakable,
    colback=pangram@boxbg, colframe=pangram@boxbg,
    boxrule=0pt, arc=2pt,
    borderline west={3pt}{0pt}{pangram@accent},
    left=14pt, right=12pt, top=10pt, bottom=10pt,
    overlay unbroken and first={\node[anchor=north east, inner sep=0pt] at ([xshift=-10pt, yshift=-7pt]frame.north east)
      {\includegraphics[height=0.72cm]{pangram-icon.png}};},
  ]
  {\sffamily\bfseries\color{pangram@heading}Abstract}\par\vspace{0.4em}
  \small
}{%
  \end{tcolorbox}\vspace{0.6em}
}
\makeatother

\begin{document}
\maketitle
\begingroup
\renewcommand{\thefootnote}{\fnsymbol{footnote}}
\footnotetext[1]{Work done during an internship at Pangram Labs.}
\endgroup

\begin{abstract}
Web text makes up the majority of pretraining data and is increasingly AI-generated. After applying FineWeb quality filtering, we find that 27.5\% of tokens from June 2026 web data are labeled as AI-generated by Pangram, rising to 31.1\% by August. Unlike synthetic data or model-collapse setups, this \emph{wild} AI text comes from many models, is written for human readers, and arrives unlabeled in pretraining corpora.
How does AI text in the wild affect language model pretraining? 
To answer this question, we pretrain 800 language models (19.9M to 973M parameters), varying the ratio of added AI tokens to human tokens, and fit scaling laws to held-out losses on both human and AI-generated text. For data-starved models, adding AI tokens to pretraining data initially lowers loss on human text, but the benefit saturates as more are added and quickly \emph{reverses} into harm. For models trained on high budgets of human text, AI tokens raise loss almost immediately, while the same number of fresh human tokens keeps lowering it. Existing scaling laws such as 
Chinchilla~\citep{hoffmann2022chinchilla}, 
which treat AI tokens as no different from human tokens, fail to predict this behavior. We propose a new scaling law with separate benefit and harm terms that allows the value of an AI token (measured in human token equivalents) to change sign while also reducing to Chinchilla in the absence of AI text. When fit on smaller models, our scaling law predicts the effect of AI text on held-out human-text loss for models up to 3.6$\times$ larger with 41\% lower error than the best existing law over all AI ratios. It implies that training on unfiltered 
web text at August 2026's AI share (31.1\%) requires 1.6$\times$ as much compute as training on its human subset at 20 tokens per parameter, a gap that grows with the human budget.
We recommend filtering AI text when the target is human text, repeating human text before expanding the training dataset with AI-generated web text, and reporting validation loss on human and AI text separately, since at a 22.3\% AI share (our 2026 evaluation crawl), mixed validation sets hide the harm in 95.5\% of harmful runs. AI text remains valuable when the target is AI text.
We release \wildai, an 83B-token corpus with AI, topic, and format labels, all 800 models and code\codelink. 
\end{abstract}

\begin{figure}[!ht]
    \centering
    \includegraphics[width=\linewidth]{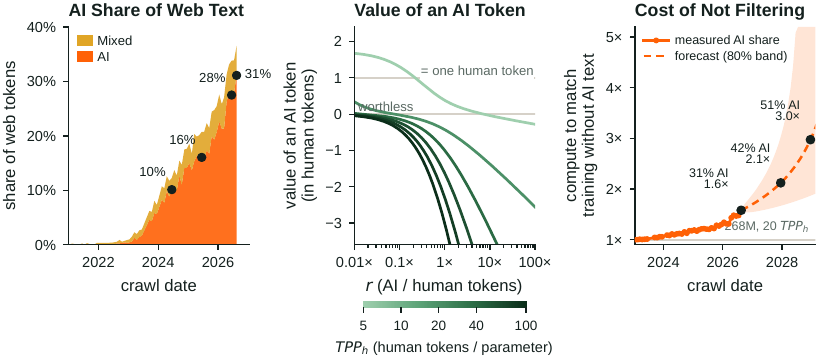}
    \caption{\textbf{AI-generated web text helps only data-starved models and costs compute otherwise.} Left: share of web tokens in documents Pangram~3.3.2 labels AI or Mixed, by month. The AI share alone is 10\% in June 2024, 16\% in June 2025, 28\% in June 2026, and 31\% in August 2026.
    Middle: our law's value of one added AI token at 268M, in human tokens, from 5 to 100 $TPP_h$: positive at 5 $TPP_h$, below zero from 20 $TPP_h$ once $r$ exceeds about 0.08. Right: compute an unfiltered crawl needs to match training on its human subset at 20 $TPP_h$ (268M) for its AI share: 1.6$\times$ at August 2026's 31\%, and 2.1$\times$ and 3.0$\times$ at the 42\% and 51\% forecasts for the ends of 2027 and 2028.
    } 
    \label{fig:law-overview}
\end{figure}

\section{Introduction}
\label{sec:intro}

AI-generated text is flooding the web.
In the June 2026 Common Crawl, Pangram labels 27.5\% of the tokens
that pass FineWeb's quality filters as AI-generated, up from 10.1\% two years earlier and rising to \textbf{\textcolor{wildaipink}{31.1\%}} in August 2026 (\autoref{fig:law-overview}). At the same time, models are trained far past the compute-optimal budget of $\sim$20 tokens per parameter~\citep{hoffmann2022chinchilla}; for example, Qwen3-32B is trained on $\sim$1,100 tokens per parameter \citep{yang2025qwen3technicalreport}. Models are also projected to exhaust nearly all public human-written text by 2032 \citep{villalobos2024runoutofdata}. Every new crawl thus forces a choice about what to do with the AI-generated documents: keep them, filter them, or find some optimal data mixture of AI and human text. 

In this paper, we measure how AI-generated web text affects pretraining and fit scaling laws that predict its impact. No previous work directly addresses this question: model-collapse studies train a model recursively on its own output \citep{shumailov2024curserecursiontraininggenerated, gerstgrasser2024is}, and synthetic-data studies add curated rephrasings designed to help performance in specific domains \citep{maini2024rephrasing, kang2025demystifying}. 
We study a third kind, which we call \emph{wild} AI text: text that language models wrote for human readers and that occurs naturally on the web, rather than being generated for training. It comes from many models, arrives in pretraining corpora unlabeled and mixed with human text, varies in quality, and makes up a rising share of the web.

We pretrain 800 models (19.9M to 973M parameters), adding up to 64 wild AI tokens per human token to fixed human corpora, and evaluate them on held-out human text (C4, FineWeb, and Paloma) as well as on AI text. AI text helps only data-starved models: it lowers loss on human text below about 10 human tokens per parameter, but from the Chinchilla-optimal 20 its benefit disappears, and larger amounts of AI tokens raise loss, while the same number of fresh human tokens still lowers it. 
Existing laws either treat AI tokens as human ones (Chinchilla), never let their value fall below zero (most repetition laws), or fit harm without separating it from benefit (most mixture laws). To address this, we propose a scaling law for wild AI-generated text 
that pairs a saturating benefit with a harm penalty that grows logarithmically and that reduces to Chinchilla without AI text. Fit on models up to 268M, it predicts models 3.6$\times$ larger with a paired error (the RMSE of the predicted change in log loss against each run's human-only control) of $0.83\times10^{-3}$ on C4, against $1.41\times10^{-3}$ for the best existing law \citep{shukor2025scaling} and $4.32\times10^{-3}$ for Chinchilla \citep{hoffmann2022chinchilla}.

Using our new law, we propose concrete recommendations for dealing with the rising share of wild AI text. Training on unfiltered web text at August 2026's AI share (31.1\%) takes 1.6$\times$ the compute of training on its human subset, rising to 3.0$\times$ by 2028 at the forecast AI share, and repeating human text beats adding AI text. Current quality filters make this worse by favoring AI text. FineWeb's pipeline \citep{penedo2024fineweb} keeps AI documents 2.3$\times$ as often as human ones, and the DCLM pipeline \citep{li2024dclm} keeps AI documents 9.8$\times$ as often. AI text is easier to predict, so a validation set with a 22.3\% AI share (our 2026 evaluation crawl) hides the harm in 95.5\% of harmful runs. For predicting AI text itself (Cosmopedia) the optimal training mix is over 90\% AI. We release \wildai, our 96M-document dataset with AI, topic, and format labels, and all 800 models.\footnote{Data and code\codelink}

\section{Measuring AI-Generated Web Data}
\label{sec:data}


In this section, we describe how we create a large-scale dataset of web data, broken down into human-written and AI-generated corpora. We find that 27.5\% of June 2026 tokens that pass the FineWeb filters are AI-generated, a share that is rising each month, that the AI share is uneven across topics and formats, and that existing pretraining quality filters (FineWeb and DCLM) favor it.

\paragraph{Collecting web data.} We use the FineWeb corpus \citep{penedo2024fineweb}, a subset of Common Crawl \citep{commoncrawl}. We extend past the June 2025 cutoff point by replicating the FineWeb filtering process on Common Crawl data from July 2025 to June 2026.\footnote{We use FineWeb v1.4.0; more details are in \S\ref{app:data}.} We collect a dataset of 96.04M documents (83.31B tokens), which we call \wildai. 

\paragraph{Detecting AI-generated content.} To understand the provenance of our data, we label each text as AI-generated or human-written. We first run EditLens Llama-3.2-3B \citep{thai2026editlens} to label vast quantities of text (more than 280B tokens, from which we draw the 83B tokens of \wildai). We then run the more accurate Pangram 3.3.2 on documents confidently labeled AI-generated and human-written as a second check, using those labels to create our final human-written corpus of 58.91M documents (42.19B tokens) and AI-generated corpus of 32.54M documents (35.32B tokens).\footnote{Because EditLens preselects likely AI documents, AI text is 42\% of \wildai's tokens, far more than on the web. \wildai is a pool for building training mixtures, not a natural sample of the web (\S\ref{app:data}).} Pangram 3.3.2 has a reported false-positive rate of 0.05\% and false-negative rate of 1.99\% \citep{glickenhaus2026pangram4}, accurate enough at this scale to separate the AI and human distributions. On pre-ChatGPT web text, Pangram 3.3.2 labels 0.062\% of 60,000 documents from 2021 crawls AI.\footnote{An upper bound on the FPR, since pre-2022 web text may itself contain some AI-generated text.} We use WebOrganizer \citep{wettig2025organize} to extract topic and format labels per document. 

\begin{figure}[t]
    \centering
    \includegraphics[width=\linewidth]{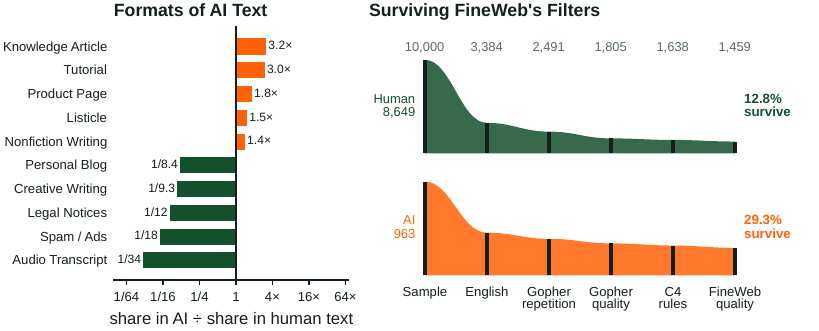}
    \caption{\textbf{AI-labeled web text leans toward a few formats and survives quality filters more often than human text.} Left: each format's share of AI-labeled tokens over its share of human-labeled tokens, January to June 2026, for the five most over- and under-represented formats. Right: share of AI- and human-labeled documents in a 2026 Common Crawl sample left after each FineWeb stage.}
    \label{fig:web-formats-filters}
\end{figure}

\paragraph{AI-generated text is changing the makeup of the internet.} To assess the increase in AI content on the web, we run Pangram on a sample of 310,000 documents, dated from January 2021 through August 2026.\footnote{5,000 documents for each month that has an associated Common Crawl.} In June of 2021, less than 0.1\% of FineWeb tokens were AI-generated. By June 2024, it was 10.1\%, 16.1\% in June 2025, and 27.5\% in June 2026 (see \autoref{fig:law-overview}, left, and \autoref{fig:web-topic-format-over-time}). The rise continues: in August 2026 the share was 31.1\%, 3.6 points above June (\S\ref{app:forecasting}). Moreover, there is a large distribution mismatch between human-written and AI-generated content that continues to grow \citep{he2026degentweblookllmdominantwebsites}. In \autoref{fig:web-formats-filters}, tutorials are 6.2\% of human-labeled tokens and 18.5\% of AI-labeled tokens, and personal blogs 8.6 and 1.0\% (see \autoref{fig:web-ai-topic-format}). 
Increasingly, the formats and topics of web data will reflect LLM outputs rather than human writing. We hypothesize that this may lead models to stray further from the human distribution on topics AI rarely writes about.

\paragraph{AI-generated data is more likely to survive quality filters.} To understand the effects of different quality filtering processes, we audit a subset of 10,000 raw Common Crawl documents and track which portion of human- and AI-written texts make it through each step of the filtration process (\autoref{fig:web-formats-filters}, see \S\ref{app:filtering}). We examine both FineWeb \citep{penedo2024fineweb} and DCLM \citep{li2024dclm} filtering processes, finding that AI-generated texts pass FineWeb's pipeline 2.3$\times$ (29.3\% vs.\ 12.8\%) and DCLM's full pipeline 9.8$\times$ (14.5\% vs.\ 1.5\%) as often as human-written documents. These filters are insufficient at filtering out AI-generated text, and in fact heavily prefer it.


\section{Background and Related Work}
\label{sec:related_work}

\paragraph{Scaling law fundamentals.} The key information needed to predict the performance of LLM training is training compute, $C$, model parameter count, $N$,
and total training tokens processed, $D$, which one can use to estimate the loss on a held-out dataset. Training compute is commonly
approximated as $C \approx 6ND$. \citet{kaplan2020scalinglawsneurallanguage} characterized these trends, and the Chinchilla scaling law proposed by \citet{hoffmann2022chinchilla} formalized them as follows: 

\begin{equation}L(N, D) = E + \frac{A}{N^{\alpha}} + \frac{B}{D^{\beta}}
\label{eq:chinchilla}
\end{equation}

\paragraph{Scaling laws for repetition.} In Chinchilla, the data term ($B$/$D^\beta$) depends only on the total token count $D$, without accounting for differences between unique tokens and repeated ones. To address this, \citet{muennighoff2023scaling} extended Chinchilla to account for repeated data by allowing for the effective value of a repeated token to be lower than a new unique token, but never below zero. \citet{qin2026bridging} propose that the effective value of a token relies not only on unique tokens but also on model size. \citet{lovelace2026prescriptive} alternatively propose that repetition is best modeled not by effective data but by adding an overfitting penalty.

\paragraph{Scaling laws for data mixtures.} Other scaling laws model pretraining data from many domains or languages. Most of this body of work focuses on optimizing pretraining data mixtures \citep{jain2024scaling, ye2025data, shukor2025scaling, hamidieh2025domainaware, sedova2026scalinglawsmixturepretraining}. Others model multilingual mixtures \citep{he2025scaling, longpre2026atlas}. 

\paragraph{Model collapse.} 
\citet{shumailov2024curserecursiontraininggenerated} study degradation under recursive training on model-generated data. Subsequent work examines how retaining original data or generating synthetic data from each successive model generation changes this behavior \citep{gerstgrasser2024is, seddik2024how}, and how collapse rates depend on the recursive estimation setting \citep{suresh2024ratemodelcollapserecursive}. \citet{schaeffer2025positionmodelcollapsedoes} show that loss increases, divergent error, and loss of distributional tails 
are distinct outcomes, while \citet{dohmatob2024a,dohmatob2025strong} show that synthetic data can affect scaling behavior even in the non-recursive setting. \citet{kang2025demystifying} find that the benefits of synthetic rephrasings and textbooks mixed with web text depend on budget. Similarly, \citet{zhu2025how} show worsened performance on Paloma as synthetic text replaces natural text at a fixed token budget. 
\citet{kazdan2025collapse} show that the benefit of synthetic data depends on total data budget.


\paragraph{Wild AI text is neither synthetic nor surrogate data.} Synthetic data is generated on purpose and curated to help, speeding pretraining 3--10$\times$ \citep{maini2024rephrasing, kang2025demystifying}, although replacing natural text with it can still hurt \citep{zhu2025how}. Model collapse trains models on their own outputs, a setting that is not as harmful once human data is kept \citep{shumailov2024curserecursiontraininggenerated, gerstgrasser2024is, kazdan2025collapse}. Wild AI text comes from many models, is written for readers and search engines rather than for training, and arrives mixed into human text, the setting that current pretraining pipelines actually face. Even the surrogate-data law of \citet{jain2024scaling} predicts our held-out C4 runs worse than Chinchilla (8.00 vs.\ 4.32, ours 0.83, all $\times10^{-3}$).


\section{Scaling Laws for Human-Written and AI-Generated Text}
\label{sec:scaling_laws}

Models trained on only human text follow Chinchilla scaling laws, but when trained on a portion of AI-generated text, AI tokens initially help models with low budgets of human text but harm models with high budgets. We create five criteria for a scaling law to model this phenomenon, and propose our scaling law that fits these criteria. Our law has the lowest error predicting the effects of adding AI-generated text versus existing scaling laws (0.83 vs.\ the second best of 1.41 $\times10^{-3}$ on C4). 

\subsection{Empirical Observations}
\label{subsec:empirical_observations}

We train 800 models of varying sizes, human tokens per parameter ($TPP_h$), and ratios of AI to human data $r$. We find that Chinchilla scaling laws are inadequate at predicting the behavior of AI-generated text and propose a new scaling law that models both the benefits and harms of pretraining on AI-generated web text.

\paragraph{Experimental setup.}
We train 800 models varying in size from 19.9M to 973M, with 2.9 to 87.5 $TPP_h$, and $r$ of 0 to 64. We use the nanochat architecture due to its fast training and high performance relative to model size \citep{nanochat}.\footnote{See \S\ref{app:training-config} for the architecture, training setup, and how $N$ is counted, following \citet{pearce2024reconciling}.} 
We evaluate on human-, mixed-, and AI-generated text sets. We use C4 \citep{raffel2020exploring} as our north star dataset to model human text, similar to the frontier Marin model evaluation \citep{marin2025retrospective}. We also evaluate on Paloma \citep{magnusson2024paloma}, reporting the loss macro-averaged over its 16 sources, and a held-out set of pre-2022 FineWeb data (FW22). For mixed text, we evaluate on a held-out set of 2026 FineWeb data (FW26), which contains 22.3\% AI-generated text. We also evaluate both the AI- and human-written splits of FineWeb 2026 (FW26-AI and FW26-H, respectively). To observe effects on purely AI text, we evaluate on Cosmopedia \citep[][Cosmo]{benallal2024cosmopedia_dataset}. Human text is our main target. It is the language a pretrained model is meant to model, and loss on held-out text is a standard proxy for downstream performance \citep{huang2024compression, gadre2025overtraining}.

\begin{figure}[!t]
    \centering
    \includegraphics[width=\linewidth]{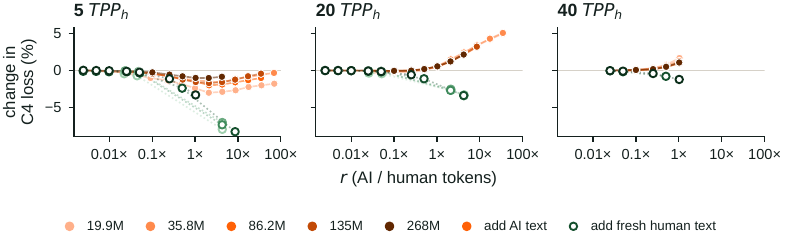}
    \caption{\textbf{Added AI text lowers loss only at small human budgets.} Change in C4 loss against each model's human-only control when AI tokens or the same number of fresh human tokens are added to a fixed human corpus, at 5, 20 and 40 $TPP_h$. }
    \label{fig:law-dose-response}
\end{figure}

\paragraph{Models trained on AI text do not follow Chinchilla scaling laws.}
We find that for models where we add more human tokens, Chinchilla scaling laws perform as expected, but adapt poorly to predicting how models with AI-generated tokens will behave. 
On the reserved sizes, Chinchilla predicts the fresh-human additions to a paired error of $1.32\times10^{-3}$ on C4 but the AI additions only to $4.32\times10^{-3}$. On models with low human token budgets, AI-generated text seems to almost always help. For Chinchilla-optimal models (20 or more $TPP_h$), AI text raises loss immediately, and more drastically at larger sizes, in line with the findings of \citet{dohmatob2025strong}. \autoref{fig:law-dose-response} shows models at different human token budgets. With 5 $TPP_h$, AI decreases loss even at large AI budgets, but at 20 $TPP_h$, the AI additions increase loss with increasing AI budgets.  


\subsection{Criteria}
\label{subsec:criteria}
Motivated by our observations, we formulate the following five criteria for a scaling law that better models the effects of adding AI-generated text:\footnote{\autoref{fig:law-benchmark} marks which benchmarked laws meet each of the five criteria.}

{\renewcommand{\labelenumi}{C\theenumi}
\begin{enumerate}
    \item Flexible token value: the value of adding an AI token must be able to change according to $N$, $TPP_h$, and $r$.
    \item Model both beneficial and harmful behavior: adding an AI token must be able to help loss in some scenarios and hurt in others.
    \item Interpretable: separate benefit and harm terms.
    \item A finite first-token value: the first AI token's value in human tokens is finite and not assumed to be 1. 
    \item Recovery of Chinchilla with no AI data: when no AI data is added, the model must recover the Chinchilla scaling laws that govern unique human tokens.
\end{enumerate}}

\subsection{Scaling Laws for AI-Generated Text}
\label{subsec:hormesis_law}

\begin{figure}[!t]
    \centering
    \includegraphics[width=\linewidth]{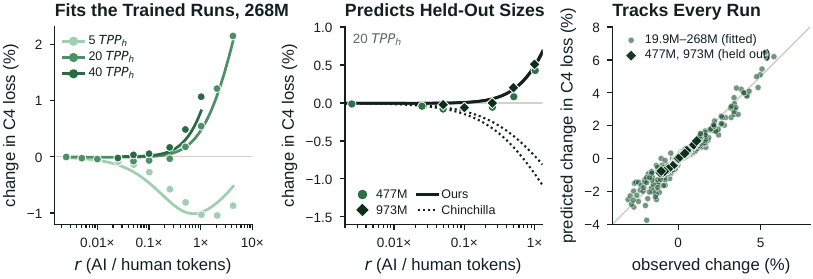}
    \caption{\textbf{Our law fits the trained models and predicts larger ones.} 
    Left: observed change in C4 loss with our law. Middle: the held-out 477M and 973M runs at 20 $TPP_h$ against our law and Chinchilla. Chinchilla predicts that AI text keeps lowering loss; the runs and our law turn upward. Right: predicted against observed change for every AI run.}
    \label{fig:law-validation}
\end{figure}

Our proposed law keeps the Chinchilla backbone of \citeauthor{hoffmann2022chinchilla} (\autoref{eq:chinchilla}), takes the saturating credit window of the compute--data (CD) law of \citeauthor{qin2026bridging} (\autoref{eq:cd_deff}) to represent the potential benefit of AI text ($D_\mathrm{eff}$), and a penalty similar to \citeauthor{lovelace2026prescriptive} (\autoref{eq:lovelace_loss}) to represent the harm ($H$) of too much AI text. The credit, harm, and net value of an AI token are depicted in \autoref{fig:law-anatomy-full}. Our law has a paired error of $0.83\times10^{-3}$ on C4's held-out sizes, lower than all other laws (\autoref{tab:law-results}). Representing human text as $D_{\mathrm{H}}$ and AI text as $D_{\mathrm{A}}$, the loss is: 

\begin{align}
{{L}}(N,D_{\mathrm{H}},D_{\mathrm{A}})
&= E + \frac{A}{N^{\alpha}} + \frac{B}{D_{\mathrm{eff}}^{\,\beta}}\,\bigl(1 + H\bigr)
\label{eq:law}
\end{align}

\paragraph{The benefits of AI tokens.} We model the saturating benefit of AI tokens as more are added. Looking to prior work on repeating tokens \citep{muennighoff2023scaling, qin2026bridging}, effective tokens increase at a decelerating rate over a window of added usefulness. 
Like both \citeauthor{muennighoff2023scaling} and \citeauthor{qin2026bridging}, we use an exponential form of credit $g(r)$, but with a free first-token parameter $\eta$.\footnote{\autoref{tab:law-ablation} in \S\ref{app:cd-fitting} lists every benefit and harm form we tried.} The benefit is as follows, where $r=D_{\mathrm{A}}/D_{\mathrm{H}}$, $R^{\star}$ is the saturation scale of the benefit, and $t=D_{\mathrm{H}}/20N$: 

\begin{align}
\quad D_{\mathrm{eff}} &= D_{\mathrm{H}}\,\bigl(1+\eta\, g(r)\bigr),
\qquad g(r)=R^{\star}\bigl(1-e^{-r/R^{\star}}\bigr),
\qquad R^{\star}=K\,t^{\rho}
\label{eq:law-credit}
\end{align}

\paragraph{The harm of AI tokens.} \citeauthor{qin2026bridging} and \citeauthor{muennighoff2023scaling} assume that repeated tokens can never \textit{hurt} performance, only help less and less. \citet{lovelace2026prescriptive} model the harm of repetition with an additive penalty, finding that the harm accelerates over epochs. We observe that the harm of adding more AI tokens \emph{decelerates} logarithmically (\autoref{fig:law-harm-per-doubling}), motivating our penalty: 

\begin{align}
\quad H &= \gamma\, t^{u}\, n^{v}\Bigl[\log(1+r)-\frac{r}{1+r}\Bigr]
\label{eq:law-harm}
\end{align}
Here $\gamma$ sets the size of the harm, $n=N/10^{8}$ is the model size, and the exponent $v$ lets the harm change with model size as $u$ lets it change with the human budget. Subtracting the AI share $r/(1+r)$ removes the initial harm of the first AI tokens. The harm grows as $r^{2}/2$ at small additions of AI and as $\log r-1$ at large ones, and the marginal harm of an AI token, proportional to $r/(1+r)^{2}$, peaks where AI and human tokens are equal ($r=1$). 

\paragraph{Relative scaling laws.} 
\citet{held2026relativescalinglawsllms} propose relative scaling laws, which measure the effect of some treatment (here, added AI text, $r > 0$) against a baseline (human text only, $r=0$). Since our scaling law reduces to Chinchilla \citep{hoffmann2022chinchilla} at $r=0$, we can predict the human-only control easily, and isolate the effect of $r$.
For every run we predict its change in log loss against its own human-only control, $\Delta=\log\widehat{\mathcal{L}}(N,D_{\mathrm{H}},D_{\mathrm{A}})-\log\widehat{\mathcal{L}}(N,D_{\mathrm{H}},0)$, and score the RMSE of predicted against observed $\Delta$ over the AI runs.\footnote{Absolute RMSE is in \autoref{tab:law-benchmark-absolute}.}

\providecommand{\lawresultstablefile}{tables/law_results_table}
\begin{table}[t]
\centering
\small
\setlength{\tabcolsep}{5pt}
\renewcommand{\arraystretch}{1.05}
\caption{Paired RMSE $\times10^{3}$ on the held-out sizes, lower is better.}
\label{tab:law-results}
\resizebox{\ifdim\width>\linewidth\linewidth\else\width\fi}{!}{%
\begin{tabular}{lrrrrr}
\toprule
Law & $k$ & C4 & FW22 & FW26-H & Paloma \\
\midrule
Chinchilla~\citep{hoffmann2022chinchilla} & 5 & \cellcolor[HTML]{CBE0D2}4.32 & \cellcolor[HTML]{CEE2D5}4.65 & \cellcolor[HTML]{D0E3D6}5.40 & \cellcolor[HTML]{B1D0BB}9.15 \\
\citet{muennighoff2023scaling} & 7 & \cellcolor[HTML]{C7DDCE}3.96 & \cellcolor[HTML]{CADFD1}4.21 & \cellcolor[HTML]{CBE0D2}4.94 & \cellcolor[HTML]{B1D0BB}9.15 \\
CD~\citep{qin2026bridging} & 8 & \cellcolor[HTML]{C4DBCB}3.68 & \cellcolor[HTML]{C2DBCA}3.63 & \cellcolor[HTML]{C1DAC9}4.17 & \cellcolor[HTML]{B1D0BB}9.15 \\
\citet{lovelace2026prescriptive} & 9 & \cellcolor[HTML]{C2DACA}3.55 & \cellcolor[HTML]{C4DBCC}3.74 & \cellcolor[HTML]{C1D9C9}4.14 & \cellcolor[HTML]{A2C6AE}8.38 \\
ATLAS~\citep{longpre2026atlas} & 6 & \cellcolor[HTML]{CCE0D2}4.36 & \cellcolor[HTML]{CFE2D5}4.72 & \cellcolor[HTML]{D1E4D7}5.51 & \cellcolor[HTML]{B1D0BB}9.15 \\
\citet{he2025scaling} & 6 & \cellcolor[HTML]{DFECE3}6.54 & \cellcolor[HTML]{DFECE3}6.55 & \cellcolor[HTML]{DCEAE0}6.62 & \cellcolor[HTML]{ADCDB8}8.96 \\
\citet{hamidieh2025domainaware} & 7 & \cellcolor[HTML]{C7DDCE}3.95 & \cellcolor[HTML]{C9DFD0}4.19 & \cellcolor[HTML]{CADFD1}4.84 & \cellcolor[HTML]{B1D0BB}9.15 \\
Shukor additive~\citep{shukor2025scaling} & 9 & \cellcolor[HTML]{E0EDE5}6.77 & \cellcolor[HTML]{E0ECE4}6.70 & \cellcolor[HTML]{D9E8DE}6.26 & \cellcolor[HTML]{C1D9C9}10.05 \\
Shukor joint~\citep{shukor2025scaling} & 13 & \cellcolor[HTML]{96BFA4}1.41 & \cellcolor[HTML]{98C1A6}1.50 & \cellcolor[HTML]{9EC4AA}2.25 & \cellcolor[HTML]{95BFA3}7.80 \\
\citet{sedova2026scalinglawsmixturepretraining} & 8 & \cellcolor[HTML]{F1F7F3}9.60 & \cellcolor[HTML]{F1F7F3}9.59 & \cellcolor[HTML]{F1F7F3}9.57 & \cellcolor[HTML]{F1F7F3}13.32 \\
\citet{jain2024scaling} & 7 & \cellcolor[HTML]{E8F2EB}8.00 & \cellcolor[HTML]{E8F2EC}8.03 & \cellcolor[HTML]{E7F1EA}8.04 & \cellcolor[HTML]{B4D2BE}9.34 \\
\textbf{Ours} & 11 & \cellcolor[HTML]{7DB08E}\textbf{0.83} & \cellcolor[HTML]{7DB08E}\textbf{0.85} & \cellcolor[HTML]{7DB08E}\textbf{1.28} & \cellcolor[HTML]{7DB08E}\textbf{6.77} \\
\bottomrule
\end{tabular}}
\end{table}

\paragraph{Benchmarking existing scaling laws.} We fit our scaling law on 726 models of 19.9M to 268M parameters, and evaluate our fit on 74 models of 477M and 973M parameters, $1.8\times$ and $3.6\times$ the largest fitted size (\autoref{tab:runs}). We benchmark existing scaling laws developed for repetition, data mixing, and surrogate data in \autoref{tab:law-results}.\footnote{\S\ref{app:published-law-implementations} defines the laws; \S\ref{app:law-benchmark-protocol} has full results.} Our law has the lowest paired error across all four human sources (0.83 $\times10^{-3}$ on C4).\footnote{On human text our law's lead over the best existing law is statistically significant over all AI ratios (\S\ref{app:significance}).} \citeauthor{shukor2025scaling}'s joint law (\autoref{eq:shukor_joint}), which treats AI and human text as two separate data mixtures, is the best of the existing laws on human text (1.41 $\times10^{-3}$ on C4) and is the best of all laws when the target text is AI-generated.\footnote{Results on mixed and AI-generated target texts are in \autoref{tab:law-results-ai-targets}.}

\section{The Effects of Training on AI Text and What to Do About It}
\label{sec:results}

We forecast that over half (50.7\%) of FineWeb tokens will be AI-generated by the end of 2028. An unfiltered crawl at August 2026's AI share (31.1\%) needs 1.6$\times$ the compute,  
and 3.0$\times$ in 2028, to match a filtered model at 20 $TPP_h$. We recommend filtering AI text from web data and, if data is short, repeating human tokens rather than adding AI tokens. AI text is worth training on when the target is AI text, and evaluations should report the two separately. 

\subsection{The Cost of Not Filtering}

\begin{figure}[!t]
    \centering
    \includegraphics[width=\linewidth]{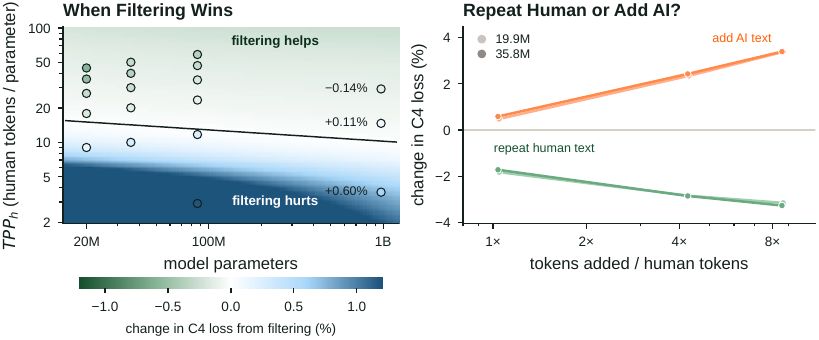}
    \caption{\textbf{Filtering AI text pays off at larger human budgets, and repeating human text beats adding AI text.} 
    Left: change in loss from removing the AI documents of a 22.3\% AI mix without replacing them, predicted by our law (color; black line: no change) and measured (circles; values for 973M). Right: change in loss when the human corpus is repeated or the same number of new AI tokens is added, for models at 20 $TPP_h$.}
    \label{fig:law-ceg}
\end{figure}

\paragraph{Filtering out AI-generated data.}
\citet{mohri2026bitterlessondatafiltering} find that with sufficient compute, it is best to not filter data at all for pretraining. To test this for AI-generated text, we train models at the 2026 natural rate of AI (22.3\%), and paired models of the same size with only the human subset of pretraining data. Despite having less data, models above about 10 to 15 $TPP_h$ (less at larger sizes) had lower loss on C4, as shown in \autoref{fig:law-ceg}. Our law gets the direction of filtering right in 18 of the 19 pairs (paired error $3.28\times10^{-3}$), while Chinchilla predicts that removing the AI text raises loss in all 13 pairs where it lowers it. We also find that the more human-written tokens a model has, the larger the benefit in loss reduction is. As most models today are trained far past 20 tokens per parameter, we recommend removing all AI-generated web text for human next-token prediction.

\paragraph{Forecasting the future AI share of web text.}
Using our historical sample of 5,000 documents per month, we use a random walk with drift \citep{hyndman2021forecasting} to forecast the future proportion of AI tokens.\footnote{More details about our forecasting methodology and predictions are in \S\ref{app:forecasting}.} Extrapolating the average monthly rise, it forecasts 42.3\% AI tokens by the end of 2027 and 50.7\% by the end of 2028.
Compute-equivalent gain (CEG) \citep{davidson2023ceg} compares the compute efficiency of any two models. Using a model trained with 22.3\% AI (the AI share of our 2026 crawl) as our reference model, our law predicts that the reference (an unfiltered model) needs 1.3$\times$ the compute to match human-only training at 268M parameters and 20 $TPP_h$ (\autoref{eq:ceg}). At August 2026's AI share (31.1\%) it needs 1.6$\times$. At the forecast AI share, by the end of 2027 a model will need 2.1$\times$ the compute to match a filtered model's performance on C4, and 3.0$\times$ by the end of 2028 (\autoref{fig:law-overview}). 

\paragraph{Is it better to repeat human data or add AI data?}
\citet{muennighoff2023scaling} and \citet{qin2026bridging} model repeated tokens as effective tokens, and \citet{lovelace2026prescriptive} with an overfitting penalty. To compare the effects of repeating human tokens and adding AI tokens, we train 19.9M and 35.8M models at 20 $TPP_h$ that add 1, 4, or 8 times the human corpus in tokens, either by repeating it or by adding AI text. For C4, loss after one added epoch was 2.3\% lower when repeating human tokens than when adding fresh AI tokens, a gap that expands to 6.3--6.4\% after eight added epochs (nine in total).\footnote{See \S\ref{app:repetition} for experimental setup details and additional results.} 

\subsection{When Is AI Text Worth Training On?}
\label{subsec:when-ai}

\begin{figure}[!t]
    \centering
    \includegraphics[width=\linewidth]{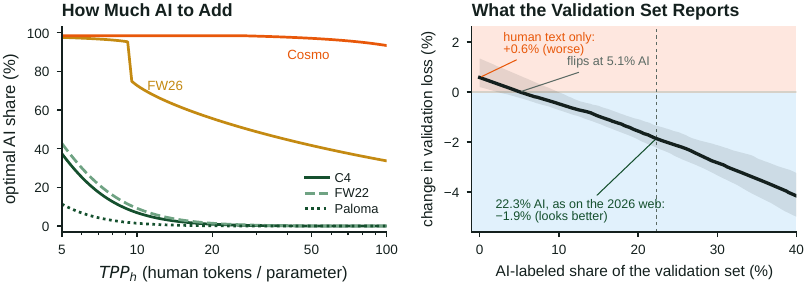}
    \caption{\textbf{Whether AI text helps depends on the evaluation set.} Left: the AI share of training tokens that minimizes our law's predicted loss with the human corpus fixed, per evaluation set: near zero for human text (C4, FW22), above 90\% for AI text (Cosmo), and falling from 98\% to 34\% for mixed FW26; the drop near 10 $TPP_h$ is where the optimum moves between two local minima of the predicted loss. Right: for the 243 AI additions that raise loss on human-labeled FW26 text, the median change in loss that a validation set reports as the AI-labeled share of its text grows. The sign flips at 5.1\% AI; at our 2026 crawl's 22.3\% the median run looks 1.9\% better.}
    \label{fig:law-when-ai}
\end{figure}

\paragraph{It helps in data-starved settings and for AI target text.} When predicting on human text, AI text helps only when human training data is scarce: the loss-minimizing AI share on C4 is 37\% at 5 $TPP_h$ but 1.0\% by 20 $TPP_h$, and FW22 and Paloma behave similarly (\autoref{fig:law-when-ai}, \S\ref{app_sec:optimal-share}). On AI-generated text, the optimal share stays above 90\% at every budget, and our law values the first AI token at more than 20 human tokens ($\eta$ in \autoref{tab:law-coefficients}). 

\paragraph{Training on AI text changes how models write.} In agentic workflows, models increasingly prompt other models and read their outputs, so a model must understand AI-written input as well as human input. Understanding such text need not mean writing it, yet training on AI text changes how a model writes: at 268M, AI-typical phrases per 1,000 words rise from 0.16 without AI text to 0.36 at $r=1$ and 0.54 at $r=4$, and at $r=2$ Pangram labels twice as many stories AI (37.2\% vs.\ 18.6\%, \autoref{tab:generation-behavior}).\footnote{See \S\ref{app:qualitative} for details on AI-typical phrases and Pangram labeling.} 

\paragraph{Treat AI and human text as separate domains, in training and in evaluation.} Mixture laws that model the two as separate domains, such as the joint law of \citet{shukor2025scaling}, are the most accurate existing laws in our benchmark, and the split could go further, since AI and human web text differ in format (tutorials are 18.5\% of AI-labeled tokens but 6.2\% of human-labeled ones). Because AI text is easier to predict, all 553 models with added AI text lower loss on AI-labeled FW26 text, while raising loss on human-labeled text in 44\% of models. At our 2026 crawl's 22.3\% AI share, a mixed validation set reports 95.5\% of these harmful runs as improvements (\autoref{fig:law-when-ai}). Tuned against C4, our law recommends 1.0\% AI at 20 $TPP_h$, and tuned against FW26, 56\% but only 2.0\% on the human subset. We recommend reporting loss on human and AI text separately.

\section{Conclusion}
\label{sec:conclusion}

Wild AI text now makes up over a quarter of filtered web crawls, and its share is rising. Across 800 pretrained models, it helps only data-starved models, increasing loss on Chinchilla-optimal models. Our law pairs a saturating benefit with a harm that grows logarithmically, reduces to Chinchilla without AI text, and predicts models 3.6$\times$ larger than those it is fit on with lower error on human text than all eleven existing laws we test. It implies that training on unfiltered web text at August 2026's AI share (31.1\%) takes 1.6$\times$ the compute of training on its human subset, rising to 3.0$\times$ by 2028, that human text should be repeated before AI text is added, and that validation loss should be reported on human and AI text separately. We release \wildai, with its AI, topic, and format labels, code, and all 800 models.

\paragraph{Limitations and future work.} We fit our law on models up to 268M parameters and test it up to 973M. Larger models memorize more of their training data \citep{carlini2023quantifying} and can be hurt more by model-generated data \citep{dohmatob2025strong}, so they may pick up the patterns of weaker LLMs more readily. They might instead learn AI and human text as related tasks that help each other, a turn the joint law of \citet{shukor2025scaling} can express. We measure next-token loss, and at our sizes AI text raises CORE scores \citep{li2024dclm} about as much as fresh human text does (\S\ref{app:downstream}), although all models are too small to have meaningful performance, and these are the tasks AI-generated text was meant to optimize. We study only pretraining, on English web text labeled by Pangram. Future work could (1) filter wild AI text, as MAI does \citep{microsoft2026mai}, and add targeted synthetic data, (2) find the domains in which AI text helps, (3) measure how the AI text already in pretraining corpora has affected the quality of current models, and, as agents read other models' outputs, (4) train models that understand AI-written input without writing like it, for example by tagging AI text \citep{caswell2019tagged} or masking its loss \citep{lin2024rho}.

\subsubsection*{Acknowledgments}

Thank you to the whole Pangram Labs team for the support and funding for this project. Thank you to the CLIP Lab at University of Maryland for their continued support and advice. Thank you to David Hall and the Marin team for allowing us to try some of our hypotheses on Marin data and providing compute for such experiments, so we could understand what is happening with wild AI text on current frontier model-sized runs (see \url{https://github.com/marin-community/marin/issues/9073}). 

\subsection*{AI Use Statement}
\label{sec:ai_use}


We used AI tools for the following tasks: 
\begin{itemize}
    \item Formulate mathematical claims: To iterate on the exact formulation of the benefit and harm terms in our scaling law, we ask coding agents to propose and test many forms of the scaling laws. 
    \item Propose or refine hypotheses: To strengthen the paper, we ask agents to suggest additional experiments to understand our scaling laws. The repetition experiment was proposed by a coding agent. 
    \item Design or provide feedback on research methodology or experiments: Coding agents were asked to provide feedback many times throughout the project, as well as design and run the training of all additive models. 
    \item Implement methods: Coding agents implemented all experimental code.
    \item Clean and reformat dataset: To extend our FineWeb dataset through June 2026, we used a coding agent to execute the FineWeb filtration process on recent Common Crawl data. 
    \item Support qualitative and thematic data analysis: We often asked coding agents to create tables and figures of results so we could interpret the results. Many of these figures and tables ended up in the final draft. 
\end{itemize}

Additional disclosures include the use of AI for the following purposes: create or modify scientific figures or images, create or edit software code,  creation of artifacts, brainstorming, sourcing/searching for information, edit a research paper to improve readability, and identify relevant literature. 

We have reviewed and validated all AI-assisted work and take responsibility for the final content of the work, including text, claims, and artifacts produced with the aid of AI.

\subsection*{Reproducibility Statement}
\label{sec:reproducability}

\textbf{Data.} \wildai is FineWeb v1.4.0 extended to June 2026 by rerunning the FineWeb pipeline on Common Crawl (\S\ref{sec:data}, \S\ref{app:data}); every document carries its EditLens and Pangram~3.3.2 provenance labels and WebOrganizer topic and format labels. We release all 83B tokens of \wildai with these document-level labels.

\textbf{Models.} All 800 models use one training recipe (\S\ref{app:training-config}) at the sizes, budgets and run grid of \autoref{tab:architecture}, \autoref{tab:budget-convention} and \autoref{tab:runs}. We release every checkpoint together with its loss on the seven evaluation sets, so that all fits, tables and figures can be regenerated on a CPU in minutes without retraining. 

\textbf{Scaling laws.} The law is \autoref{eq:law} to \autoref{eq:law-harm} with the fitted coefficients of \autoref{tab:law-coefficients}; the objective, starting points, bounds and the fixed development/reserved split are in \S\ref{app:cd-fitting}; every comparator is written in our coordinates in \S\ref{app:published-law-implementations} and scored by the same protocol (\S\ref{app:law-benchmark-protocol}); the alternative forms we tried are in \autoref{tab:law-ablation}. 

\textbf{Code.} The data pipeline, labeling, training, evaluation, law fitting and every figure and table script are released with the data and models.

\bibliographystyle{plainnat}
\bibliography{iclr2027_conference}

\appendix
\section{Data}
\label{app:data}

\paragraph{Collecting web data.}
\wildai combines two samples of English web text. The first is drawn from the 16 released FineWeb dumps, CC-MAIN-2024-10 to CC-MAIN-2025-26 \citep{penedo2024fineweb}. The second extends FineWeb to the twelve Common Crawl dumps from July 2025 to June 2026 (CC-MAIN-2025-30 to CC-MAIN-2026-25): we rerun FineWeb's DataTrove pipeline \citep{penedo2024datatrove} on the first 2.6\% of each dump's WARC files, keeping 5.9M documents (4.6B tokens). In total, \wildai has 96.04M documents and 83.31B tokens.

\paragraph{Labeling with EditLens and Pangram.}
\label{app:labeling}
EditLens (Llama-3.2-3B) first sorts documents into four buckets from human-written to AI-generated, averaging over up to three evenly spaced 512-token windows. We take its most human bucket as human candidates and its two most AI buckets as AI candidates, dropping the bucket between them. Pangram then labels every candidate: we run the Pangram~3.3.2 model on up to three disjoint 512-token windows per document (the first, middle and last) and call a document Human or AI only when every window agrees, and Mixed otherwise. On 2,000 documents from June 2026, these labels match those of the Pangram~3.3.2 API on 96.7\%; only 4 documents flip between Human and AI (the others involve mixed). The filter audit of \S\ref{app:filtering} is labeled the same way. We use the Pangram label to make the training pool, comprising 58.91M human documents (42.19B tokens) and 32.54M AI documents (35.32B tokens), with the 4.59M Mixed documents left out. FW26 holds the Human- and AI-labeled documents of a random sample of our January to June 2026 extension: its Mixed documents (7\% of its tokens) are left out, and 22.3\% is the AI share of the remaining tokens (20.1\% of the tokens scored, since the evaluation drops the part of a document beyond one 2,048-token sequence and AI documents are longer). The filtering experiments of \S\ref{sec:results} train on our human and AI pools mixed at that share. The monthly AI shares of \S\ref{sec:data} come from a separate sample of 5,000 documents drawn at random from each of the 62 crawl months between January 2021 and August 2026 and labeled through the Pangram~3.3.2 API, taken from FineWeb through June 2025 and from randomly drawn WARC files of each crawl after (\S\ref{app:forecasting}).

\paragraph{False positives on pre-ChatGPT text.} Web text crawled before ChatGPT's release in November 2022 \citep{openai2022chatgpt} should contain almost no AI text, so the share Pangram labels AI there bounds its false-positive rate on web documents. Of the 60,000 documents the monthly sample draws from 2021 (5,000 for each of the twelve calendar months spanned by the nine 2021 crawls), Pangram~3.3.2 labels 37 AI (0.062\%, 95\% interval 0.045 to 0.085\%; 0.04\% of tokens) and 25 Mixed, in line with its reported false-positive rate of 0.05\%. Some of these documents may be genuinely machine-generated, since earlier language models already wrote web text, so the false-positive rate on web text is at most this.

\begin{figure}[!t]
    \centering
    \includegraphics[width=\linewidth]{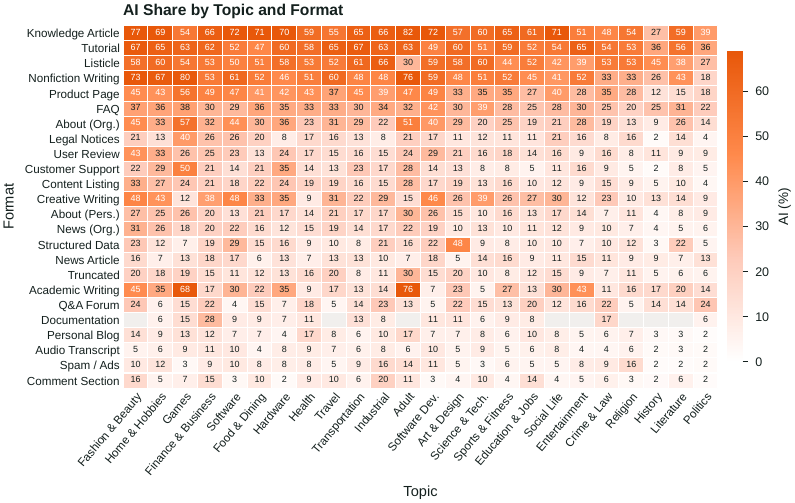}
    \caption{Knowledge articles, tutorials and listicles are the most AI-labeled formats in nearly every topic. Share of documents Pangram~3.3.2 labels AI in each WebOrganizer topic and format of our training pool documents.}
    \label{fig:web-ai-topic-format}
\end{figure}

\begin{figure}[!t]
    \centering
    \includegraphics[width=\linewidth]{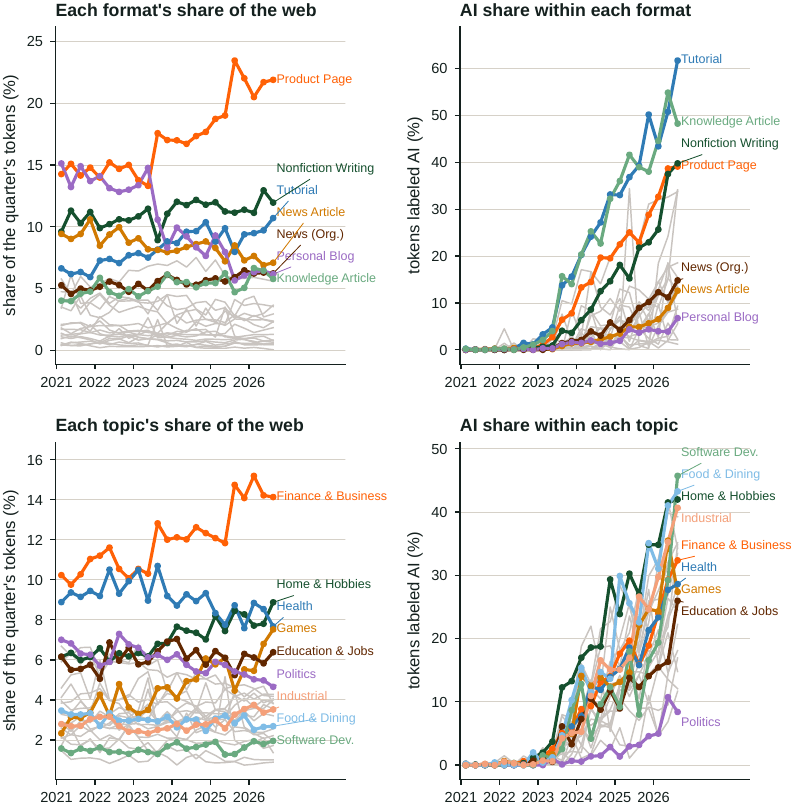}
    \caption{The web shifts toward the formats and topics AI writes most. We plot January 2021 to August 2026, Pangram~3.3.2 and WebOrganizer labels by calendar quarter. Each format's and topic's share of web tokens and the share of each format's and topic's tokens labeled AI.}
    \label{fig:web-topic-format-over-time}
\end{figure}

\paragraph{AI formats and topics.}
We find that the makeup of the web is changing over time. Product pages grow from 14 to 22\% of tokens while personal blogs fall from 15 to 6\%. In 2026, about half of tutorial and knowledge-article tokens are AI-labeled (51 and 49\%), against 9\% of news and 5\% of personal-blog tokens, as shown in \autoref{fig:web-topic-format-over-time}. In our training pool, which was assembled to supply AI text (about 34\% AI documents, against 8 to 13\% on the web in 2024 and 2025), knowledge articles (60\% of documents), tutorials (59\%) and listicles (53\%) lead the formats, against about 6\% for comment sections, audio transcripts and spam. Fashion and beauty, home and hobbies, games and finance (40 to 46\%) lead the topics, and politics (12\%) trails, as shown in \autoref{fig:web-ai-topic-format}.

\subsection{Effects of Quality Filtering}
\label{app:filtering}

\begin{figure}[!t]
    \centering
    \includegraphics[width=\linewidth]{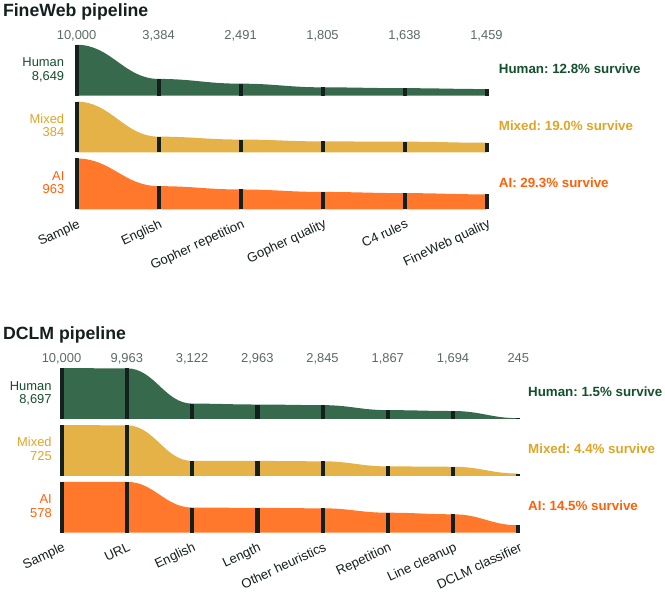}
    \caption{\textbf{DCLM's quality classifier keeps AI-labeled documents far more often than human ones.} Survival of a 10,000-document 2026 Common Crawl sample through each stage of the FineWeb (top) and DCLM (bottom) pipelines, split by Pangram~3.3.2 label.}
    \label{fig:web-filter-survival}
\end{figure}

To understand the effects of quality filtration systems on AI-generated text in relation to human text, we sample 10,000 documents from 2026 Common Crawl. We replicate the FineWeb and DCLM filtration processes, and trace the proportion of AI-generated text that passes through each step in relation to the proportion of human text that survives. The filtration process is depicted in \autoref{fig:web-filter-survival}. Each pipeline runs on its own extraction of a 10,000-document sample from the same 32 WARC files of CC-MAIN-2026-30: FineWeb's on text extracted with Trafilatura, as FineWeb does, and DCLM's on text extracted with Resiliparse, as DCLM does, so the two samples hold different numbers of documents per label.

\paragraph{AI text survives both pipelines more often.} FineWeb keeps 29.3\% of AI-labeled documents (282 of 963) and 12.8\% of human-labeled ones (1,104 of 8,649), 2.3$\times$ as often (95\% interval 2.1 to 2.6$\times$). AI documents pass its English filter more often (45\% against 33\%), and among English documents its quality heuristics keep 64\% of AI documents and 39\% of human ones. DCLM keeps 14.5\% of AI documents (84 of 578) and 1.5\% of human ones (129 of 8,697), 9.8$\times$ as often (7.5 to 12.7$\times$). The gap opens at its learned classifier, which keeps 40\% of the AI documents that reach it and 9.5\% of the human ones. AI documents survive more often within the same format, so the gap is not only a matter of format mix: among English documents, DCLM keeps 53\% of AI-labeled knowledge articles against 17\% of human ones, and 49\% of AI tutorials against 20\%.

\paragraph{Which human documents do filters remove?} Most removed human-labeled documents fail the English filters or are extraction debris (93\% of truncated pages, 88\% of content listings and 85\% of structured data are removed after FineWeb's English filter), but DCLM's learned classifier also removes most human prose that reaches it, including 111 of 117 news articles and 106 of 117 personal blogs.

\subsection{Forecasting the AI Share of the Web}
\label{app:forecasting}

\begin{figure}[!t]
    \centering
    \includegraphics[width=0.72\linewidth]{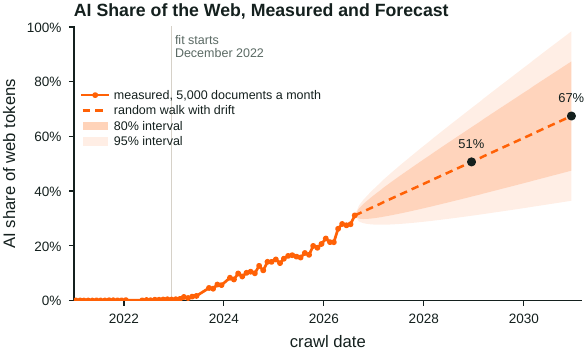}
    \caption{The AI share of web tokens, measured each crawl month and forecast to December 2030 by a random walk with drift fitted to December 2022 through August 2026, with 80 and 95\% prediction intervals.}
    \label{fig:web-ai-share-forecast}
\end{figure}

Every month of the series covers its whole crawl. Up to June 2025 the 5,000 documents come from FineWeb, whose dumps span every WARC file of a crawl and which we read in a seeded random shard order. From July 2025 on, they come from 32 WARC files drawn at random from each crawl and run through the FineWeb filters. To forecast our future rates of AI, we take a sample of 10,000 documents from July and August 2026, past our data cutoff dates, as an evaluation sample. We find the rate of AI tokens in the samples was 27.9\% for July and 31.1\% for August. We compare several forecasting methods on held-out months, including ones that respect the bounds of a share (the lowest proportion of the web AI tokens can hold is 0, and the highest 100). We try ARIMA \citep{box1970time}, random walks with drift \citep{hyndman2021forecasting}, ordinary damped Holt \citep{gardner1985forecasting}, and logit-damped methods (a damped trend on the logit of the share; \citealp{hyndman2021forecasting}) and choose a random walk with drift because it gave the most accurate predictions on months since ChatGPT \citep{openai2022chatgpt} was introduced in late November 2022. The July and August samples come from crawls after our June 2026 data cutoff (CC-MAIN-2026-30 and CC-MAIN-2026-34). They first served as a check on forecasts fit through June; the final forecasts refit the chosen random walk with drift through August 2026. We estimate 50.7\% of text will be AI by the end of 2028, shown in \autoref{fig:web-ai-share-forecast} and \autoref{tab:ai-share-forecast}. 

\begin{table}[t]
\centering
\small
\setlength{\tabcolsep}{6pt}
\caption{Forecast AI share of web tokens each December. The intervals are Student-$t$ prediction intervals that include the uncertainty of the drift.}
\label{tab:ai-share-forecast}
\begin{tabular}{lrrr}
\toprule
December & Forecast & 80\% interval & 95\% interval \\
\midrule
2026 & 33.9\% & 30.0 to 37.8\% & 27.8 to 40.0\% \\
2027 & 42.3\% & 33.5 to 51.1\% & 28.7 to 55.9\% \\
2028 & 50.7\% & 38.0 to 63.4\% & 31.0 to 70.4\% \\
2029 & 59.1\% & 42.7 to 75.5\% & 33.7 to 84.5\% \\
2030 & 67.5\% & 47.5 to 87.4\% & 36.5 to 98.4\% \\
\bottomrule
\end{tabular}
\end{table}

\section{Training Configurations}
\label{app:training-config}

\paragraph{Architecture details.} We report architecture details in \autoref{tab:architecture}. All 800 models share one nanochat training recipe \citep{nanochat}. The input embedding and output head are untied. $N$ counts the input embedding but not nanochat's value-embedding tables, which are also lookups (16.8M parameters at 19.9M, 709M at 973M). The tokenizer is nanochat's BPE with 32,768 tokens, and sequences hold 2,048 tokens. Muon \citep{jordan2024muon} updates the transformer matrices (learning rate 0.02) and AdamW \citep{loshchilov2019decoupled} the embeddings, output head and scalars. The learning rate warms up over 40 steps, stays constant, and decays linearly over the last 65\% of steps to 5\% of its peak. A step holds 262,144 tokens at 19.9M and 35.8M, 524,288 from 86.2M to 268M, and 1,048,576 at 477M and 973M. The schedule spans all of a run's tokens, so a run with added tokens trains for more steps than its control and decays later. Human and AI documents are interleaved in a fixed hashed order, and every AI or fresh-human addition trains on its control's human documents, in the same order and with the same seed. 703 of the 800 runs use one seed; the other 97, all at 19.9M to 268M, are replicates with other seeds (46 cells trained with two, three or five seeds). Training uses bf16 with FP8 matrix multiplies on eight H100 GPUs. We score the final checkpoint in bits per byte on 15.7M tokens each of C4 and FW22, 12.1M of FW26, 9.6M of FW26-H, 2.2M of FW26-AI and 18.5M of Cosmopedia, each document scored at most once, and the validation split of Paloma's 16 sources. \autoref{tab:run-inventory} lists all 800 runs, and \autoref{tab:runs} shows the 365 in the matched-budget grid. The 38 filtering models of \autoref{fig:law-ceg} (left) and the six repetition models of \S\ref{app:repetition} are not among the 800 and enter no fit.

\begin{table}[t]
\centering
\small
\setlength{\tabcolsep}{6pt}
\caption{The seven model sizes. All use the nanochat architecture. $N$ in the scaling laws counts the input embedding; the last column subtracts it. Reserved sizes ($\dagger$) enter no fit.}
\label{tab:architecture}
\begin{tabular}{lrrrrr}
\toprule
Model & Layers & Width & Heads & $N$ (with input embedding) & Without \\
\midrule
19.9M & 4 & 256 & 2 & 19.9M & 11.5M \\
35.8M & 6 & 384 & 3 & 35.8M & 23.2M \\
86.2M & 9 & 640 & 5 & 86.2M & 65.2M \\
135M & 12 & 768 & 6 & 135.3M & 110.1M \\
268M & 16 & 1024 & 8 & 268.4M & 234.9M \\
477M$^{\dagger}$ & 20 & 1280 & 10 & 477.1M & 435.2M \\
973M$^{\dagger}$ & 26 & 1664 & 13 & 972.9M & 918.4M \\
\bottomrule
\end{tabular}
\end{table}

\begin{table}[t]
\centering
\small
\setlength{\tabcolsep}{4.5pt}
\caption{Every model in the paper. The 800 models of the scaling-law cohort by size and arm: human-only controls, AI additions and fresh-human additions, each addition trained on exactly its control's human documents. Sizes marked $\dagger$ are held out from every fit. $TPP_h$: range over the controls. \emph{In \autoref{tab:runs}}: runs in the matched-budget grid, controls included; the others belong to control groups at per-size budgets from earlier run generations and enter every fit and score in the same way. \emph{Other seeds}: runs with a seed other than 1337. Below the rule, the 44 filtering and repetition models, which enter no fit.}
\label{tab:run-inventory}
\begin{tabular}{lcrrrrrr}
\toprule
 & & & \multicolumn{2}{c}{Additions} & & & \\
\cmidrule(lr){4-5}
Model & $TPP_h$ & Controls & AI & Human & Total & In \autoref{tab:runs} & Other seeds \\
\midrule
19.9M & 2.9--87.5 & 20 & 113 & 20 & 153 & 61 & 25 \\
35.8M & 3.2--87.5 & 16 & 96 & 16 & 128 & 60 & 3 \\
86.2M & 3.8--87.5 & 17 & 88 & 8 & 113 & 54 & 7 \\
135M & 4.1--87.5 & 18 & 110 & 45 & 173 & 73 & 30 \\
268M & 4.4--87.5 & 17 & 95 & 47 & 159 & 69 & 32 \\
\cmidrule(lr){1-8}
\emph{Fitted} & & 88 & 502 & 136 & 726 & 317 & 97 \\
\midrule
477M$^{\dagger}$ & 4.6--73.0 & 8 & 42 & 12 & 62 & 36 & 0 \\
973M$^{\dagger}$ & 4.7--37.8 & 3 & 9 & 0 & 12 & 12 & 0 \\
\cmidrule(lr){1-8}
\emph{Held out} & & 11 & 51 & 12 & 74 & 48 & 0 \\
\midrule
\textbf{All} & & \textbf{99} & \textbf{553} & \textbf{148} & \textbf{800} & \textbf{365} & \textbf{97} \\
\midrule
\multicolumn{8}{l}{\emph{Outside the cohort, never fitted}} \\
\multicolumn{5}{l}{Filtering (\autoref{fig:law-ceg}, left): 22.3\% AI mix, and AI removed} & 38 & & \\
\multicolumn{5}{l}{Repetition (\S\ref{app:repetition}): human corpus repeated} & 6 & & \\
\bottomrule
\end{tabular}
\end{table}

\begin{table}[t]
\centering
\small
\setlength{\tabcolsep}{5pt}
\caption{The training grid. Each cell is one human-only control plus the number of AI-addition runs and, after the plus sign, fresh-human-addition runs on the same human corpus. Sizes marked $\dagger$, below the rule, are reserved: every law is scored on them and none is fitted on them.}
\label{tab:runs}
\begin{tabular}{lcccccc|c}
\hline
Model $\backslash$ $TPP_h$ & 5 & 20 & 40 & 55 & 75 & 90 & Other \\
\hline
19.9M & 12\,+\,3 & 11\,+\,4 & 5 & 9\,+\,1 & 5 & 5 & 14: 66\,+\,12 \\
35.8M & 12\,+\,3 & 11\,+\,4 & 5 & 8\,+\,1 & 5 & 5 & 10: 50\,+\,8 \\
86.2M & 11\,+\,3 & 9\,+\,3 & 5 & 7 & 5 & 5 & 11: 46\,+\,2 \\
135M & 14\,+\,7 & 12\,+\,7 & 6\,+\,3 & 5\,+\,1 & 5\,+\,1 & 5\,+\,1 & 12: 63\,+\,25 \\
268M & 11\,+\,7 & 11\,+\,6 & 6\,+\,4 & 5\,+\,1 & 5\,+\,1 & 5\,+\,1 & 11: 52\,+\,27 \\
\hline
477M$^{\dagger}$ & 7\,+\,3 & 7\,+\,3 & 5 & 3 & 3 & -- & 3: 17\,+\,6 \\
973M$^{\dagger}$ & 0 & 5 & 4 & -- & -- & -- & -- \\
\hline
\end{tabular}
\end{table}

\paragraph{Parameter count and human budget.} Following \citet{pearce2024reconciling} (see also \citealp{weng2026scaling}), $N$ counts the input embedding, as nanochat reports model sizes.\footnote{\url{https://github.com/karpathy/nanochat/discussions/420}} This choice also fits our laws better at the smallest sizes, where the two counts differ most. Figures label each human budget by its $TPP_h$ rounded to the nearest 5, and \autoref{tab:budget-convention} gives the exact values. 
\begin{table}[t]
\centering
\small
\setlength{\tabcolsep}{6pt}
\caption{$TPP_h$ of the aligned budgets. 477M has no 90 budget; 973M has no 55, 75 or 90 budget.}
\label{tab:budget-convention}
\begin{tabular}{lrrrrrr}
\toprule
Label in figures ($TPP_h$) & 5 & 20 & 40 & 55 & 75 & 90 \\
Actual $D_{\mathrm H}/N$ & 4.7 & 18.9 & 37.8 & 54.7 & 73.0 & 87.5 \\
\bottomrule
\end{tabular}
\end{table}

\section{Additional Scaling Law Background}
\label{app_sec:scaling_law_background}

\subsection{Scaling Laws for Repetition}
\citet{lovelace2026prescriptive} propose an alternative to effective data, arguing that there is an interaction between model size and repetition that the saturation term did not capture in their experiments. They instead keep the total token count $D=U_D(1+R_D)$ in the data term and add an overfitting penalty. The penalty is fit by coefficient $P$, which grows with repetition and model capacity relative to the number of unique tokens, allowing for validation loss to increase if the cost of repetition outweighs the benefits. Their simplest form is:

\begin{equation}
    L(N,U_D,R_D)
    =
    E + \frac{A}{N^\alpha}
      + \frac{B}{[U_D(1+R_D)]^\beta}
      + \underbrace{
          P\,R_D\,\frac{N}{U_D}
        }_{\text{overfitting penalty}}
\label{eq:lovelace_loss}
\end{equation}

\citet{muennighoff2023scaling} extended Chinchilla to account for repeated data using an effective token count, $D_{\mathrm{eff}}$, and an effective parameter count, $N_{\mathrm{eff}}$, written as follows:

\begin{equation}
    L(N,U_D,R_D)
    =
    E + \frac{A}{N_{\mathrm{eff}}^\alpha}
      + \frac{B}{D_{\mathrm{eff}}^\beta}
\label{eq:muennighoff_loss}
\end{equation}

Let $U_D$ denote the number of unique training tokens
and $R_D$ the number of additional epochs beyond the first.
Then $D=U_D(1+R_D)$. $R_D^\star$ is a fitted constant controlling
the saturation of repetition. Then effective data is defined as:

\begin{equation}
    D_{\mathrm{eff}}
    =
    U_D
    + \underbrace{
        U_D R_D^\star
        \left(1-e^{-R_D/R_D^\star}\right)
      }_{\text{discounted credit for repeated tokens}},
    \qquad
    R_D = \frac{D}{U_D}-1
\label{eq:muennighoff_effective_tokens}
\end{equation}

In the benchmark (\autoref{tab:law-results}) the human tokens are the unique data, $U_D=D_{\mathrm{H}}$, and the added AI tokens play the part of repeated data, $R_D=D_{\mathrm{A}}/D_{\mathrm{H}}$.

\citet{qin2026bridging} model the dependence of data saturation on the amount of unique tokens and model size. They extend \citet{muennighoff2023scaling}, proposing that the effective value of a token depends on $N$, fresh tokens per parameter $\mathrm{TPP}=U_D/N$, and $R_D$. For paraphrased data, $R_D=(D-U_D)/U_D$ measures added derived tokens relative to fresh tokens. Their Compute-Data (CD) law is:

\begin{equation}
    L(N,U_D,R_D)
    =
    E + \frac{A}{N^\alpha}
      + \frac{B}{D_{\mathrm{eff}}^\beta}
      \label{eq:cd_loss}
\end{equation}

Unlike the constant $R_D^\star$ in \citet{muennighoff2023scaling}, CD's saturation scale varies with $N$ and $\mathrm{TPP}$. The effective token count and saturation scale are detailed below:

\begin{equation}
    D_{\mathrm{eff}}
    =
    U_D\left[
      1+R_D^\star
      \left(1-e^{-R_D/R_D^\star}\right)
    \right],
    \qquad
    R_D^\star
    =
    K\,\mathrm{TPP}^{\rho}N^{\sigma}
\label{eq:cd_deff}
\end{equation}

\subsection{Scaling Laws for Data Mixtures}
Closest to our setting, \citet{shukor2025scaling} fit mixture laws for a second data source. To model the effects of AI-generated text, we adapt their law with human share $h=1/(1+r)$ and AI share $f=r/(1+r)$ of the total token count $D=D_{\mathrm{H}}+D_{\mathrm{A}}$. Their additive form keeps the Chinchilla constants and adds a term in the shares:
\begin{equation}
L(N,D,f) = E + \frac{A}{N^{\alpha}} + \frac{B}{D^{\beta}} + \frac{1}{c_{\mathrm{H}}h^{q_{\mathrm{H}}} + c_{\mathrm{A}}f^{q_{\mathrm{A}}}},
\label{eq:shukor_additive}
\end{equation}

They also propose a joint form that lets the shares rescale the model and data amplitudes, as follows:

\begin{equation}
L(N,D,f) = E + \frac{(a_{\mathrm{H}}h + a_{\mathrm{A}}f)^{p_A}}{N^{\alpha}} + \frac{(b_{\mathrm{H}}h + b_{\mathrm{A}}f)^{p_B}}{D^{\beta}} + \frac{1}{c_{\mathrm{H}}h^{q_{\mathrm{H}}} + c_{\mathrm{A}}f^{q_{\mathrm{A}}}},
\label{eq:shukor_joint}
\end{equation}

We benchmark the joint form as the strongest published comparator (\autoref{tab:law-results}).

\subsection{Other Benchmarked Scaling Laws}
\label{app:published-law-implementations}
We compare published functional forms, specialized to two data sources where needed, with coefficients fitted on our experiments. We use $n=N/10^8$, $d=D_{\mathrm H}/10^9$, $r=D_{\mathrm A}/D_{\mathrm H}$, $f=r/(1+r)$, and $t=D_{\mathrm H}/(20N)$. The remaining published forms are written below in our coordinates with $M=E+A/n^\alpha$, $T=d(1+r)$, $h=1-f$, and $q>0$.
\begin{align}
L_{\mathrm{ATLAS}}&=M+B[d(1+qr)]^{-\beta}\\
L_{\mathrm{He}}&=(M+B T^{-\beta})h^{-\zeta}\\
L_{\mathrm{Hamidieh}}&=M+B T^{-\beta}
 \exp[-a_{\mathrm H}h\log D_{\mathrm H}-a_{\mathrm A}f\log D_{\mathrm A}]\\
L_{\mathrm{Jain}}&=M+B\left[\frac{1+qr}{d(1+r)^2}\right]^\beta+\gamma f^2\\
L_{\mathrm{Sedova}}&=M+B n^c[d(1+qr)]^{-\beta}+\gamma f
\label{eq:other-local-transfers}
\end{align}

\section{Scaling-Law Fitting and Model Selection}
\label{app:cd-fitting}

\begin{figure}[!t]
    \centering
    \includegraphics[width=\linewidth]{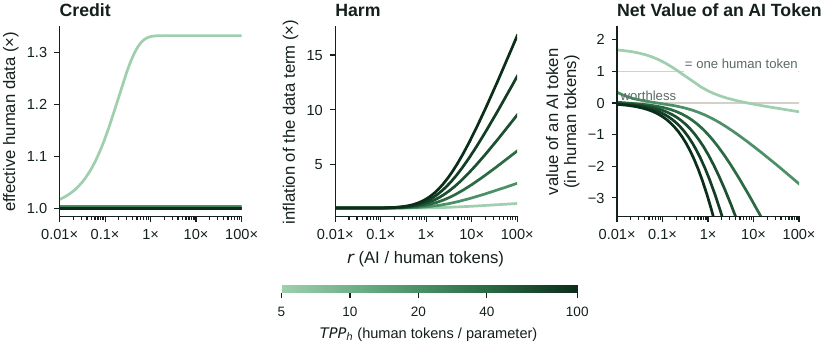}
    \caption{\textbf{The credit and harm terms of our law.} The two terms of \autoref{eq:law} and their net effect on C4 at 268M, for human budgets from 5 to 100 $TPP_h$.}
    \label{fig:law-anatomy-full}
\end{figure}

\paragraph{Fitting our scaling laws.} Every law, ours and each comparator, is fitted on the same 726 models at the five development sizes (19.9M to 268M). The 74 models at 477M and 973M are held out. The human-only controls contribute their own residual, and every addition run (AI or fresh human) contributes the residual of its change in log loss against its own control, so that the fit is anchored on the paired contrasts of \S\ref{sec:scaling_laws}. Coefficients minimize a Huber loss on these residuals with bounded least squares, from a Chinchilla fit on the human-only runs as the starting point. \autoref{tab:law-coefficients} lists the coefficients of our law fitted separately on each evaluation target, and \autoref{fig:law-anatomy-full} draws its credit and harm terms on C4.

\begin{table}[t]
\centering
\small
\setlength{\tabcolsep}{4pt}
\caption{Coefficients of \autoref{eq:law} fitted separately on each evaluation target, on the same 726 models (19.9M to 268M) as \autoref{tab:law-results}.}
\label{tab:law-coefficients}
\begin{tabular}{lrrrrrrr}
\toprule
 & \multicolumn{4}{c}{\textbf{Human text}} & \multicolumn{1}{c}{Mixed} & \multicolumn{2}{c}{AI text} \\
\cmidrule(lr){2-5}\cmidrule(lr){6-6}\cmidrule(lr){7-8}
Coefficient & C4 & FW22 & FW26-H & Paloma & FW26 & FW26-AI & Cosmo \\
\midrule
$E$ & 0.7115 & 0.6683 & 0.6001 & 0.5003 & 0.5565 & 0.1827 & 0.2886 \\
$A$ & 0.2548 & 0.2759 & 0.2728 & 0.9020 & 0.2655 & 0.1725 & 0.2152 \\
$\alpha$ & 0.3027 & 0.2900 & 0.2919 & 0.1198 & 0.2993 & 0.4269 & 0.3714 \\
$B$ & 0.08480 & 0.09414 & 0.09822 & 0.02689 & 0.1063 & 0.4261 & 0.2761 \\
$\beta$ & 0.4722 & 0.4485 & 0.4350 & 0.9614 & 0.4149 & 0.1209 & 0.1815 \\
$\eta$ & 1.729 & 1.629 & 1.678 & 2.886 & 3.292 & 25.52 & 22.51 \\
$K$ & 0.001997 & 0.003274 & 0.004705 & 0.0003355 & 26.72 & 48.85 & 36.86 \\
$\rho$ & -3.293 & -3.098 & -2.912 & -3.629 & -0.3194 & -0.5681 & -0.5227 \\
$\gamma$ & 0.4053 & 0.3398 & 0.3083 & 5.000 & 1.624 & 0.05901 & 0.1042 \\
$u$ & 1.204 & 1.197 & 1.167 & 1.395 & 0.3881 & 0.6943 & 0.6465 \\
$v$ & 0.4422 & 0.4209 & 0.4155 & 1.042 & -0.1266 & -0.5363 & -0.5801 \\
\bottomrule
\end{tabular}
\end{table}

\subsection{Choosing the Scaling Law}

\paragraph{Choosing the data benefit.}

\begin{table}[t]
\centering
\small
\setlength{\tabcolsep}{4pt}
\renewcommand{\arraystretch}{1.1}
\caption{Our law on C4 with the harm or benefit changed one at a time: paired RMSE $\times10^{3}$, lower is better, scored on the held-out sizes.}
\label{tab:law-ablation}
\resizebox{\ifdim\width>\linewidth\linewidth\else\width\fi}{!}{%
\begin{tabular}{lrr}
\toprule
Form & $k$ & 477M and 973M \\
\midrule
\multicolumn{3}{l}{\emph{The credit AI tokens earn (harm as in our law)}} \\
\quad No credit: AI tokens add no data & 8 & 3.29 \\
\quad Linear, $\eta r$ (no saturation) & 9 & 1.87 \\
\quad Exponential window, $\eta=1$ (CD's credit) & 10 & 0.97 \\
\quad Rational window, free $\eta$ & 11 & 0.98 \\
\quad Tanh window, free $\eta$ & 11 & \textbf{0.83} \\
\quad Exponential window, no budget dependence ($\rho=0$) & 10 & 1.71 \\
\quad Exponential window, also size-dependent ($R^\star=K t^\rho n^\sigma$) & 12 & 1.41 \\
\quad \textbf{Exponential window, free $\eta$, $R^\star=K t^\rho$ (ours)} & 11 & \textbf{0.83} \\
\midrule
\multicolumn{3}{l}{\emph{The harm AI tokens do (credit as in our law)}} \\
\quad No harm (credit only) & 8 & 3.27 \\
\quad $\gamma\,[\log(1+r)-r/(1+r)]$ on the data term, no budget or size dependence & 9 & 2.83 \\
\quad $\gamma t^{u}[\log(1+r)-r/(1+r)]$ on the data term, no size dependence & 10 & 2.36 \\
\quad $\gamma n^{v}[\log(1+r)-r/(1+r)]$ on the data term, no budget dependence & 10 & 2.16 \\
\quad $\gamma t^{u}n^{v}[\log(1+r)-r/(1+r)]$ added to the loss, not tied to the data term & 11 & \textbf{0.80} \\
\quad $\gamma t^{u}n^{v}\log(1+r)$ on the data term (harm starts at the first AI token) & 11 & 1.15 \\
\quad Box--Cox of $\log(1+r)$ with free curvature $\lambda$, on the data term & 12 & 1.13 \\
\quad Power law $\gamma r^{p} t^{u} n^{v}$ added to the loss \citep{lovelace2026prescriptive} & 12 & 1.55 \\
\quad \textbf{$\gamma t^{u}n^{v}[\log(1+r)-r/(1+r)]$ on the data term (ours)} & 11 & 0.83 \\
\bottomrule
\end{tabular}}
\end{table}

\begin{figure}[!t]
    \centering
    \includegraphics[width=\linewidth]{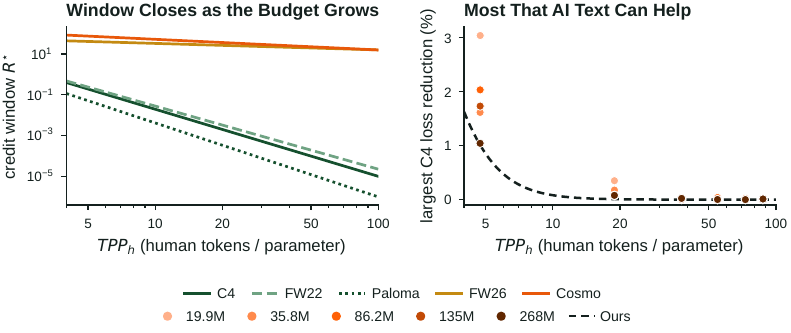}
    \caption{\textbf{The window in which AI text helps closes as the human budget grows.} Left: the fitted credit window $R^{\star}=K\,t^{\rho}$, the rate at which the benefit of AI text saturates, against the human budget for each evaluation set; it is orders of magnitude wider for the mixed and AI-text sets (FW26, Cosmo) than for human text. Right: the largest reduction in C4 loss that added AI text achieves in each fitted group and under our law: up to 3.0\% at 5 $TPP_h$, under 0.4\% at 20 $TPP_h$ and none beyond.}
    \label{fig:style-cd-window}
\end{figure}

\autoref{tab:law-ablation} changes the credit term of our law one form at a time, fitting each on C4 at the development sizes and scoring it on the held-out sizes. We choose an exponential term to make the fewest changes from \citet{muennighoff2023scaling} and \citet{qin2026bridging}, just adding a free parameter $\eta$ that can control the size of the benefit. On the held-out sizes it scores 0.83 $\times10^{-3}$, the same as a tanh window. \autoref{fig:style-cd-window} shows how the fitted window saturates over different human token budgets and the largest amount of performance AI can help a model at different human token budgets.

\begin{figure}[!t]
    \centering
    \includegraphics[width=0.6\linewidth]{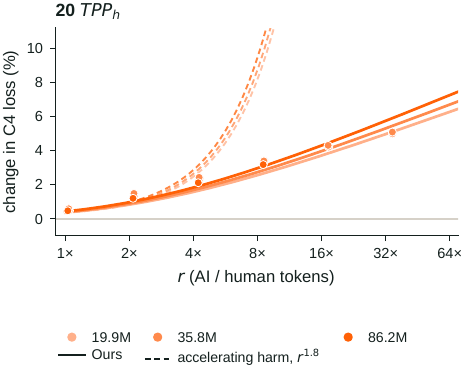}
    \caption{\textbf{The harm of AI text decelerates.} Change in C4 loss against $r$ from 1 to 64: each doubling of AI text adds the same increase in loss. Dashed: an accelerating power-law penalty in the style of \citet{lovelace2026prescriptive} (exponent 1.8, for illustration)}.
    \label{fig:law-harm-per-doubling}
\end{figure}

\paragraph{Choosing the penalty.} The power-law harm of \citet{lovelace2026prescriptive} would show the AI-generated harm accelerating, whereas we empirically show the increase in loss slowing down, as shown in \autoref{fig:law-harm-per-doubling}; fitted to our runs, a power-law penalty takes an exponent below one (0.61 on C4), so it too decelerates. We try many penalties and choose the logarithm minus the AI share, $\log(1+r)-r/(1+r)$; every form tried is in \autoref{tab:law-ablation}. Adding the same harm to the loss instead of the data term scores 0.80 on C4 against our 0.83 $\times10^{-3}$, inside bootstrap noise (95\% interval of the gap $-0.07$ to $+0.01$); the two forms are near-reparametrizations of each other and give the same cost of not filtering at 8B (1.58 and 1.59$\times$). We keep the harm on the data term because it keeps the whole effect of AI text inside the data term, so that the law remains Chinchilla in an effective token count.

\subsection{Law Fits Across Model Sizes and Budgets}

\begin{figure}[!t]
    \centering
    \includegraphics[width=\linewidth]{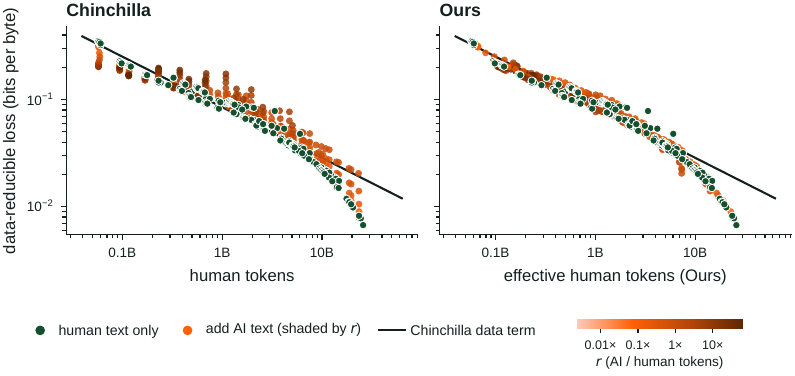}
    \caption{Effective human tokens place the AI runs on the Chinchilla data term. Counting human tokens only, AI runs scatter off the Chinchilla data term traced by the human-only runs; counted as effective tokens, they collapse onto it.}
    \label{fig:style-cd-effective-data-collapse}
\end{figure}

\begin{figure}[!t]
    \centering
    \includegraphics[width=\linewidth]{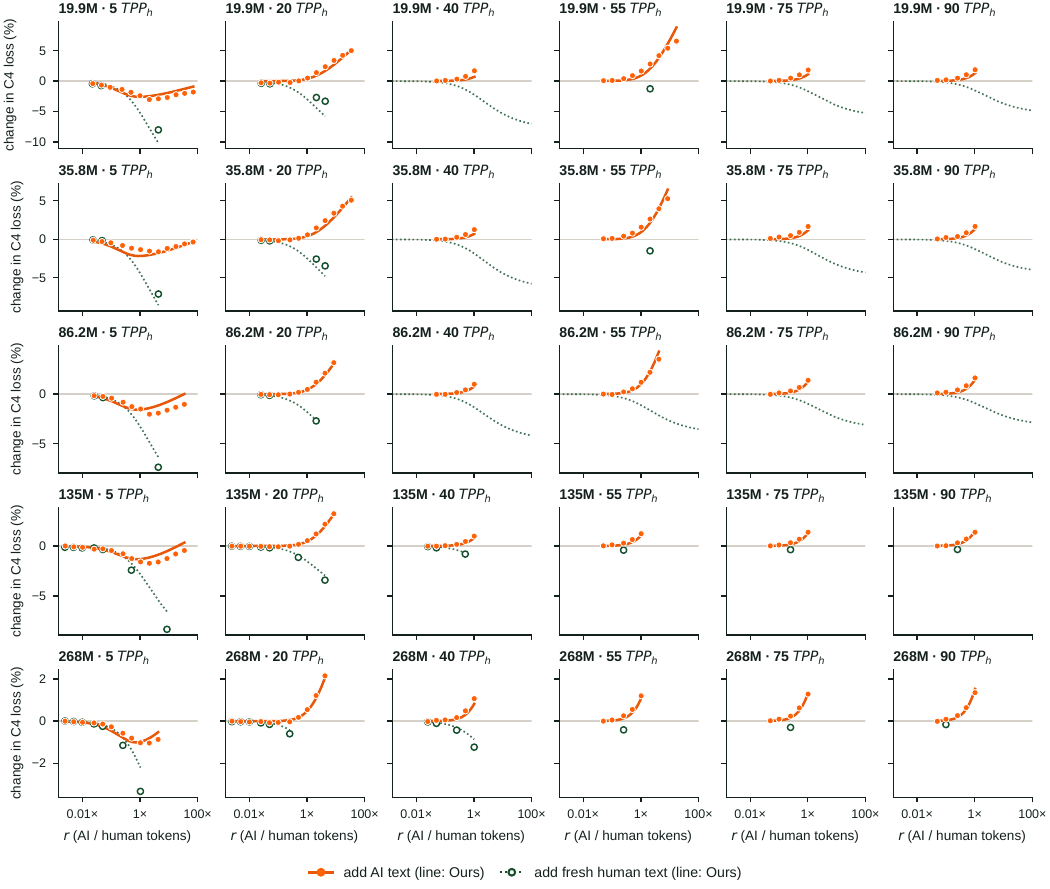}
    \caption{Our law across every fitted size and budget. Change in C4 loss against $r$ for every fitted control group with our law's curves for added AI text and added fresh human text.}
    \label{fig:law-dose-ladder-all}
\end{figure}

\begin{figure}[!t]
    \centering
    \includegraphics[width=0.8\linewidth]{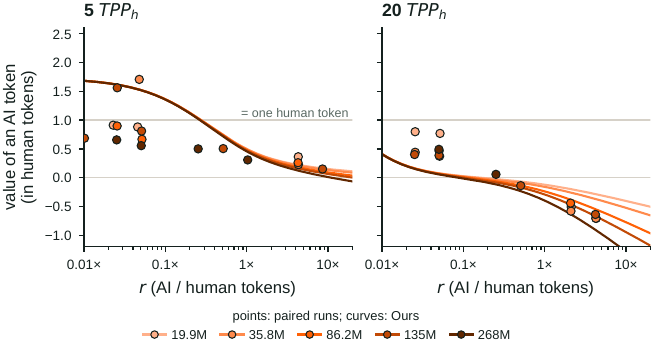}
    \caption{\textbf{Our law's value of an AI token against measured values.} Value of an added AI token in human tokens under our law and measured from paired AI and fresh-human additions to the same control, at 5 and 20 $TPP_h$.}
    \label{fig:law-token-value-calibration}
\end{figure}

\autoref{fig:style-cd-effective-data-collapse} compares the Chinchilla fit to ours. \autoref{fig:law-dose-ladder-all} shows all $TPP_h$ and size fits on C4. 
At low $TPP_h$ added AI text first lowers loss and then raises it; from about 20 $TPP_h$ it raises loss from the first tokens, and the harm grows roughly with $\log r$. \autoref{fig:law-token-value-calibration} compares the value of an AI token under our law with the value measured from paired AI and fresh-human runs.

\section{Benchmarking Other Scaling Laws}
\label{app:law-benchmark-protocol}

\begin{figure}[!t]
    \centering
    \includegraphics[width=\linewidth]{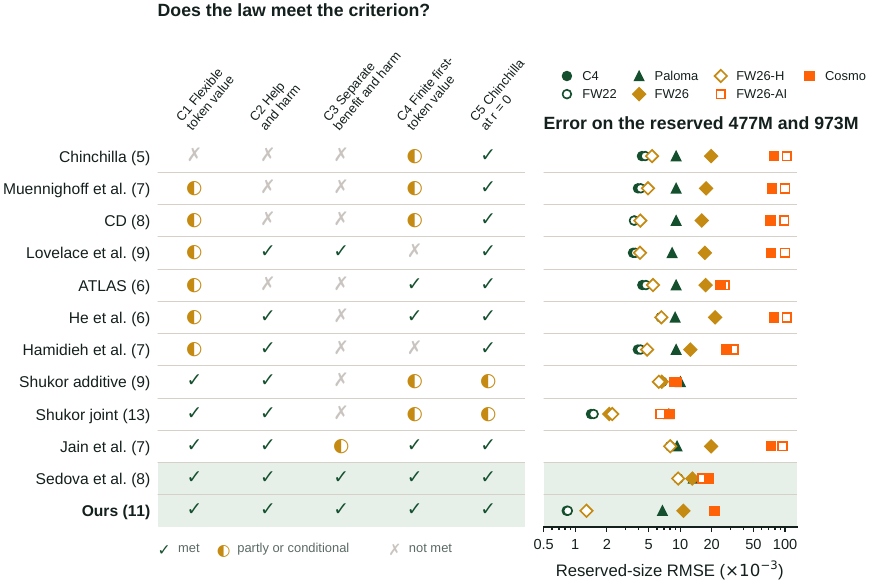}
    \caption{\textbf{Only our law meets all five criteria at low error.} Left: whether each benchmarked law meets the criteria of \S\ref{subsec:criteria}: always (check), partly or only for some coefficient values (half circle), or for no coefficient values (cross). For criterion C4 (a finite first-token value), a half circle marks a value that is finite but fixed, as in Chinchilla, where every AI token is worth exactly one human token, or finite only when a fitted exponent reaches one. Right: each law's paired RMSE on the held-out runs for every evaluation set, the numbers of \autoref{tab:law-results} and \autoref{tab:law-results-ai-targets}. The law of \citet{sedova2026scalinglawsmixturepretraining} also meets every criterion, at 11.6 times our error on C4.}
    \label{fig:law-benchmark}
\end{figure}

We benchmark 11 other scaling laws, applying laws meant for repetition, data mixing, and surrogate data to see how AI-generated text most closely acts. The requirements each law meets and their performance on held-out runs are depicted in \autoref{fig:law-benchmark}. Additional analyses benchmark laws on absolute loss instead of relative (\autoref{tab:law-benchmark-absolute}), on AI-generated target text (\autoref{tab:law-results-ai-targets}), and on low ratios of AI-generated tokens (\autoref{tab:law-results-lowdose}). On Paloma, five comparators (Chinchilla, \citeauthor{muennighoff2023scaling}, CD, ATLAS and \citeauthor{hamidieh2025domainaware}) fit a data term that is negligible at the held-out token counts ($\beta$ at its upper bound of 2.5, or for CD a credit window at its lower bound, $R^{\star}<0.003$). They predict no change from added AI text ($|\Delta|\le2.1\times10^{-5}$), so all five score the error of predicting zero change, 9.15 (7.75 at $r<1$). Refitting Chinchilla with $\beta$ allowed up to 10 gives the same error. On absolute error the ranking changes: CD and ATLAS predict the loss of the human-text sets best (12.51 and 13.65 on C4, against 14.09 for ours, all $\times10^{-3}$), while ours is best on Paloma and on the AI-text sets (\autoref{tab:law-benchmark-absolute}). Because the paired errors are much smaller, absolute error mostly measures how well a law predicts the loss of the human-only models at the held-out sizes. The paired error removes that level and isolates the effect of AI text, which is what the laws are meant to capture, so it is our main metric.

\begin{table}[t]
\centering
\small
\setlength{\tabcolsep}{4pt}
\renewcommand{\arraystretch}{1.1}
\caption{\autoref{tab:law-results} for the mixed and AI-text targets: paired RMSE $\times10^{3}$ on the held-out sizes, lower is better. On FW26-AI and Cosmo, the exponent of the human-share multiplier of \citet{he2025scaling} sits at its bound of zero, where the law is exactly Chinchilla: the multiplier can only raise the loss as AI text is added, and added AI text lowers the loss on these targets.}
\label{tab:law-results-ai-targets}
\resizebox{\ifdim\width>\linewidth\linewidth\else\width\fi}{!}{%
\begin{tabular}{lrrrr}
\toprule
 & & \multicolumn{3}{c}{477M and 973M} \\
\cmidrule(lr){3-5}
Law & $k$ & FW26 & FW26-AI & Cosmo \\
\midrule
\multicolumn{5}{l}{\emph{General}} \\
\quad Chinchilla~\citep{hoffmann2022chinchilla} & 5 & \cellcolor[HTML]{F6F5F4}19.74 & \cellcolor[HTML]{F8F7F6}103.86 & \cellcolor[HTML]{F8F7F6}78.76 \\
\multicolumn{5}{l}{\emph{Repetition and paraphrase}} \\
\quad \citet{muennighoff2023scaling} & 7 & \cellcolor[HTML]{F4F3F1}17.67 & \cellcolor[HTML]{F7F6F5}99.83 & \cellcolor[HTML]{F7F6F5}75.30 \\
\quad CD~\citep{qin2026bridging} & 8 & \cellcolor[HTML]{F2F1EF}16.06 & \cellcolor[HTML]{F7F6F5}98.08 & \cellcolor[HTML]{F6F5F4}72.83 \\
\quad \citet{lovelace2026prescriptive} & 9 & \cellcolor[HTML]{F3F2F1}17.18 & \cellcolor[HTML]{F7F6F5}99.75 & \cellcolor[HTML]{F6F5F4}73.07 \\
\multicolumn{5}{l}{\emph{Data mixtures}} \\
\quad ATLAS~\citep{longpre2026atlas} & 6 & \cellcolor[HTML]{F4F2F1}17.47 & \cellcolor[HTML]{E1DEDC}26.56 & \cellcolor[HTML]{E0DDDB}24.29 \\
\quad \citet{he2025scaling} & 6 & \cellcolor[HTML]{F8F7F6}21.55 & \cellcolor[HTML]{F8F7F6}103.86 & \cellcolor[HTML]{F8F7F6}78.76 \\
\quad \citet{hamidieh2025domainaware} & 7 & \cellcolor[HTML]{EDEBEA}12.50 & \cellcolor[HTML]{E4E2E0}32.33 & \cellcolor[HTML]{E2E0DE}27.61 \\
\quad Shukor additive~\citep{shukor2025scaling} & 9 & \cellcolor[HTML]{E0DEDB}6.67 & \cellcolor[HTML]{CFCBC8}9.19 & \cellcolor[HTML]{CBC7C3}8.85 \\
\quad Shukor joint~\citep{shukor2025scaling} & 13 & \cellcolor[HTML]{C9C5C1}\textbf{2.11} & \cellcolor[HTML]{C9C5C1}\textbf{6.45} & \cellcolor[HTML]{C9C5C1}\textbf{7.98} \\
\quad \citet{sedova2026scalinglawsmixturepretraining} & 8 & \cellcolor[HTML]{EEECEB}13.07 & \cellcolor[HTML]{D9D6D3}16.25 & \cellcolor[HTML]{DBD8D5}18.81 \\
\multicolumn{5}{l}{\emph{AI data}} \\
\quad \citet{jain2024scaling} & 7 & \cellcolor[HTML]{F6F5F4}19.73 & \cellcolor[HTML]{F6F5F4}94.23 & \cellcolor[HTML]{F7F5F4}73.46 \\
\quad \textbf{Ours} & 11 & \cellcolor[HTML]{EAE8E6}10.74 & \cellcolor[HTML]{DDDBD8}21.32 & \cellcolor[HTML]{DDDAD7}20.92 \\
\bottomrule
\end{tabular}}
\end{table}

\begin{table}[t]
\centering
\small
\setlength{\tabcolsep}{4pt}
\renewcommand{\arraystretch}{1.1}
\caption{Absolute BPB RMSE $\times10^{3}$ on the held-out sizes, lower is better: the error in the predicted loss level itself rather than in the paired change against the human-only control, so the ranking can differ.}
\label{tab:law-benchmark-absolute}
\resizebox{\ifdim\width>\linewidth\linewidth\else\width\fi}{!}{%
\begin{tabular}{lrrrrr}
\toprule
\multicolumn{6}{l}{\textbf{Human-text targets}} \\
 & & \multicolumn{4}{c}{477M and 973M} \\
\cmidrule(lr){3-6}
Law & $k$ & C4 & FW22 & FW26-H & Paloma \\
\midrule
\multicolumn{6}{l}{\emph{General}} \\
\quad Chinchilla~\citep{hoffmann2022chinchilla} & 5 & \cellcolor[HTML]{F1F7F3}26.93 & \cellcolor[HTML]{F1F7F3}28.41 & \cellcolor[HTML]{F1F7F3}29.74 & \cellcolor[HTML]{AFCFB9}30.91 \\
\multicolumn{6}{l}{\emph{Repetition and paraphrase}} \\
\quad \citet{muennighoff2023scaling} & 7 & \cellcolor[HTML]{D0E3D7}21.70 & \cellcolor[HTML]{CDE1D4}22.24 & \cellcolor[HTML]{C8DECF}22.76 & \cellcolor[HTML]{BFD9C8}31.66 \\
\quad CD~\citep{qin2026bridging} & 8 & \cellcolor[HTML]{7DB08E}\textbf{12.51} & \cellcolor[HTML]{7DB08E}\textbf{12.86} & \cellcolor[HTML]{7DB08E}\textbf{13.91} & \cellcolor[HTML]{95BFA3}29.76 \\
\quad \citet{lovelace2026prescriptive} & 9 & \cellcolor[HTML]{B7D3C0}18.35 & \cellcolor[HTML]{B3D1BD}18.57 & \cellcolor[HTML]{B1D0BB}19.57 & \cellcolor[HTML]{9FC5AB}30.19 \\
\multicolumn{6}{l}{\emph{Data mixtures}} \\
\quad ATLAS~\citep{longpre2026atlas} & 6 & \cellcolor[HTML]{8AB899}13.65 & \cellcolor[HTML]{8AB89A}14.08 & \cellcolor[HTML]{8AB899}15.17 & \cellcolor[HTML]{AECEB9}30.88 \\
\quad \citet{he2025scaling} & 6 & \cellcolor[HTML]{A2C7AE}16.00 & \cellcolor[HTML]{9DC4AA}15.98 & \cellcolor[HTML]{9DC3AA}17.12 & \cellcolor[HTML]{95BFA3}29.77 \\
\quad \citet{hamidieh2025domainaware} & 7 & \cellcolor[HTML]{9BC2A8}15.25 & \cellcolor[HTML]{9FC5AC}16.24 & \cellcolor[HTML]{A1C6AE}17.65 & \cellcolor[HTML]{B1D0BC}31.02 \\
\quad Shukor additive~\citep{shukor2025scaling} & 9 & \cellcolor[HTML]{A0C6AD}15.79 & \cellcolor[HTML]{9AC1A7}15.62 & \cellcolor[HTML]{96BFA4}16.40 & \cellcolor[HTML]{ADCDB8}30.82 \\
\quad Shukor joint~\citep{shukor2025scaling} & 13 & \cellcolor[HTML]{9CC3A9}15.36 & \cellcolor[HTML]{99C1A7}15.61 & \cellcolor[HTML]{92BDA0}15.95 & \cellcolor[HTML]{87B697}29.17 \\
\quad \citet{sedova2026scalinglawsmixturepretraining} & 8 & \cellcolor[HTML]{AECEB9}17.32 & \cellcolor[HTML]{A3C7AF}16.65 & \cellcolor[HTML]{A1C6AE}17.63 & \cellcolor[HTML]{F1F7F3}34.05 \\
\multicolumn{6}{l}{\emph{AI data}} \\
\quad \citet{jain2024scaling} & 7 & \cellcolor[HTML]{B6D3BF}18.21 & \cellcolor[HTML]{AECEB9}17.97 & \cellcolor[HTML]{AACBB5}18.64 & \cellcolor[HTML]{AECEB9}30.87 \\
\quad \textbf{Ours} & 11 & \cellcolor[HTML]{8FBB9E}14.09 & \cellcolor[HTML]{8DBA9C}14.31 & \cellcolor[HTML]{8CB99B}15.33 & \cellcolor[HTML]{7DB08E}\textbf{28.74} \\
\bottomrule
\end{tabular}}
\par\vspace{8pt}
\resizebox{\ifdim\width>\linewidth\linewidth\else\width\fi}{!}{%
\begin{tabular}{lrrrr}
\toprule
\multicolumn{5}{l}{\textbf{Mixed and AI-text targets}} \\
 & & \multicolumn{3}{c}{477M and 973M} \\
\cmidrule(lr){3-5}
Law & $k$ & FW26 & FW26-AI & Cosmo \\
\midrule
\multicolumn{5}{l}{\emph{General}} \\
\quad Chinchilla~\citep{hoffmann2022chinchilla} & 5 & \cellcolor[HTML]{F6F5F4}31.36 & \cellcolor[HTML]{F6F5F4}61.27 & \cellcolor[HTML]{F6F5F3}43.27 \\
\multicolumn{5}{l}{\emph{Repetition and paraphrase}} \\
\quad \citet{muennighoff2023scaling} & 7 & \cellcolor[HTML]{E1DFDC}22.49 & \cellcolor[HTML]{F7F6F5}63.51 & \cellcolor[HTML]{F8F7F6}47.41 \\
\quad CD~\citep{qin2026bridging} & 8 & \cellcolor[HTML]{D4D1CD}18.27 & \cellcolor[HTML]{F8F7F6}67.76 & \cellcolor[HTML]{F7F6F5}46.40 \\
\quad \citet{lovelace2026prescriptive} & 9 & \cellcolor[HTML]{DEDBD9}21.42 & \cellcolor[HTML]{F7F6F5}65.63 & \cellcolor[HTML]{F8F7F6}47.72 \\
\multicolumn{5}{l}{\emph{Data mixtures}} \\
\quad ATLAS~\citep{longpre2026atlas} & 6 & \cellcolor[HTML]{F8F7F6}32.39 & \cellcolor[HTML]{D0CDC9}9.80 & \cellcolor[HTML]{D5D2CF}10.86 \\
\quad \citet{he2025scaling} & 6 & \cellcolor[HTML]{E3E1DF}23.31 & \cellcolor[HTML]{F6F5F4}61.27 & \cellcolor[HTML]{F6F5F3}43.27 \\
\quad \citet{hamidieh2025domainaware} & 7 & \cellcolor[HTML]{DFDDDA}21.84 & \cellcolor[HTML]{D6D3D0}12.88 & \cellcolor[HTML]{D8D5D2}12.15 \\
\quad Shukor additive~\citep{shukor2025scaling} & 9 & \cellcolor[HTML]{CECAC7}16.63 & \cellcolor[HTML]{DDDBD8}18.60 & \cellcolor[HTML]{DEDBD9}15.79 \\
\quad Shukor joint~\citep{shukor2025scaling} & 13 & \cellcolor[HTML]{D1CECA}17.51 & \cellcolor[HTML]{DBD9D6}16.86 & \cellcolor[HTML]{DCDAD7}14.81 \\
\quad \citet{sedova2026scalinglawsmixturepretraining} & 8 & \cellcolor[HTML]{C9C5C1}\textbf{15.34} & \cellcolor[HTML]{CDCAC6}8.42 & \cellcolor[HTML]{C9C5C1}6.65 \\
\multicolumn{5}{l}{\emph{AI data}} \\
\quad \citet{jain2024scaling} & 7 & \cellcolor[HTML]{E3E0DE}23.07 & \cellcolor[HTML]{EDEBEA}39.82 & \cellcolor[HTML]{EEECEB}31.39 \\
\quad \textbf{Ours} & 11 & \cellcolor[HTML]{D1CDCA}17.40 & \cellcolor[HTML]{C9C5C1}\textbf{6.85} & \cellcolor[HTML]{C9C5C1}\textbf{6.54} \\
\bottomrule
\end{tabular}}
\end{table}

\begin{table}[t]
\centering
\small
\setlength{\tabcolsep}{4pt}
\renewcommand{\arraystretch}{1.1}
\caption{The paired errors of \autoref{tab:law-results} restricted to the held-out runs with $r<1$, the range a web crawl can reach (our 2026 crawl is $r=0.29$; an even split of AI and human text is $r=1$).}
\label{tab:law-results-lowdose}
\resizebox{\ifdim\width>\linewidth\linewidth\else\width\fi}{!}{%
\begin{tabular}{lrrrrr}
\toprule
 & & \multicolumn{4}{c}{477M and 973M} \\
\cmidrule(lr){3-6}
Law & $k$ & C4 & FW22 & FW26-H & Paloma \\
\midrule
\multicolumn{6}{l}{\emph{General}} \\
\quad Chinchilla~\citep{hoffmann2022chinchilla} & 5 & \cellcolor[HTML]{C4DCCC}2.93 & \cellcolor[HTML]{C6DDCE}3.18 & \cellcolor[HTML]{C7DDCF}3.82 & \cellcolor[HTML]{A5C9B1}7.75 \\
\multicolumn{6}{l}{\emph{Repetition and paraphrase}} \\
\quad \citet{muennighoff2023scaling} & 7 & \cellcolor[HTML]{BCD7C5}2.47 & \cellcolor[HTML]{BDD7C6}2.63 & \cellcolor[HTML]{BCD7C5}3.21 & \cellcolor[HTML]{A5C9B1}7.75 \\
\quad CD~\citep{qin2026bridging} & 8 & \cellcolor[HTML]{B7D4C1}2.22 & \cellcolor[HTML]{B3D1BD}2.12 & \cellcolor[HTML]{AFCFBA}2.60 & \cellcolor[HTML]{A5C9B1}7.75 \\
\quad \citet{lovelace2026prescriptive} & 9 & \cellcolor[HTML]{CDE1D4}3.50 & \cellcolor[HTML]{CEE1D4}3.69 & \cellcolor[HTML]{CBE0D2}4.07 & \cellcolor[HTML]{A9CBB5}7.88 \\
\multicolumn{6}{l}{\emph{Data mixtures}} \\
\quad ATLAS~\citep{longpre2026atlas} & 6 & \cellcolor[HTML]{C5DCCD}2.96 & \cellcolor[HTML]{C7DDCF}3.24 & \cellcolor[HTML]{C9DED0}3.92 & \cellcolor[HTML]{A5C9B1}7.75 \\
\quad \citet{he2025scaling} & 6 & \cellcolor[HTML]{DBEAE0}4.78 & \cellcolor[HTML]{DBEAE0}4.83 & \cellcolor[HTML]{D8E8DD}5.03 & \cellcolor[HTML]{A4C8B0}7.70 \\
\quad \citet{hamidieh2025domainaware} & 7 & \cellcolor[HTML]{C1DACA}2.75 & \cellcolor[HTML]{C1DAC9}2.86 & \cellcolor[HTML]{BFD9C8}3.38 & \cellcolor[HTML]{A5C9B1}7.75 \\
\quad Shukor additive~\citep{shukor2025scaling} & 9 & \cellcolor[HTML]{D7E7DC}4.34 & \cellcolor[HTML]{D5E6DB}4.29 & \cellcolor[HTML]{C8DED0}3.90 & \cellcolor[HTML]{B2D0BC}8.16 \\
\quad Shukor joint~\citep{shukor2025scaling} & 13 & \cellcolor[HTML]{91BC9F}0.98 & \cellcolor[HTML]{8EBA9D}1.01 & \cellcolor[HTML]{9AC2A7}1.82 & \cellcolor[HTML]{8FBB9E}7.06 \\
\quad \citet{sedova2026scalinglawsmixturepretraining} & 8 & \cellcolor[HTML]{F1F7F3}7.54 & \cellcolor[HTML]{F1F7F3}7.58 & \cellcolor[HTML]{F1F7F3}7.59 & \cellcolor[HTML]{F1F7F3}10.66 \\
\multicolumn{6}{l}{\emph{AI data}} \\
\quad \citet{jain2024scaling} & 7 & \cellcolor[HTML]{CDE1D4}3.51 & \cellcolor[HTML]{CADFD1}3.43 & \cellcolor[HTML]{BDD7C6}3.27 & \cellcolor[HTML]{96BFA4}7.27 \\
\quad \textbf{Ours} & 11 & \cellcolor[HTML]{7DB08E}\textbf{0.64} & \cellcolor[HTML]{7DB08E}\textbf{0.71} & \cellcolor[HTML]{7DB08E}\textbf{1.14} & \cellcolor[HTML]{7DB08E}\textbf{6.54} \\
\bottomrule
\end{tabular}}
\end{table}

\subsection{Statistical Significance}
\label{app:significance}
We resample the 51 held-out AI runs by their 10 control groups (10,000 draws), hold every fit fixed, and recompute each law's paired error (\autoref{tab:law-significance}). Over all AI ratios our law has lower error than the joint law of \citet{shukor2025scaling} in 99.99\% of draws on C4 (gap 0.58 $\times10^{-3}$, 95\% interval 0.32 to 0.84) and in every draw on FW22 and FW26-H, and lower error than each other comparator in every draw. On Paloma our law beats the joint law in 98.7\% of draws but is the best of all twelve laws in only 89.4\%.

\begin{table}[t]
\centering
\footnotesize
\setlength{\tabcolsep}{3pt}
\caption{Our law's lead holds over all AI ratios on human text, but not at $r<1$ on C4 and FW22. Paired RMSE $\times10^{3}$ on the held-out runs for our law and the joint law of \citet{shukor2025scaling}, with 95\% intervals for ours and for the gap from 10,000 bootstrap draws that resample the 51 held-out AI runs (43 with $r<1$) by their 10 control groups, every fit held fixed. The last two columns give the share of draws in which our law has lower error than the joint law, and than all eleven comparators.}
\label{tab:law-significance}
\begin{tabular}{llccccc}
\toprule
 & & & & & \multicolumn{2}{c}{Draws where ours is (\%)} \\
\cmidrule(lr){6-7}
Target & AI runs & Ours & Joint law & Joint $-$ ours & below joint & best of 12 \\
\midrule
C4 & all & 0.83 [0.64, 0.99] & 1.41 & +0.58 [+0.32, +0.84] & 99.99 & 99.99 \\
 & $r<1$ & 0.64 [0.53, 0.77] & 0.98 & +0.34 [-0.03, +0.65] & 96.5 & 96.5 \\
\addlinespace[2pt]
FW22 & all & 0.85 [0.74, 0.96] & 1.50 & +0.65 [+0.32, +0.94] & 100 & 100 \\
 & $r<1$ & 0.71 [0.56, 0.87] & 1.01 & +0.30 [-0.10, +0.65] & 92.6 & 92.6 \\
\addlinespace[2pt]
FW26-H & all & 1.28 [0.97, 1.64] & 2.25 & +0.97 [+0.58, +1.35] & 100 & 100 \\
 & $r<1$ & 1.14 [0.75, 1.55] & 1.82 & +0.69 [+0.23, +1.06] & 99.8 & 99.8 \\
\addlinespace[2pt]
Paloma & all & 6.77 [3.31, 9.57] & 7.80 & +1.03 [+0.14, +1.65] & 98.7 & 89.4 \\
 & $r<1$ & 6.54 [3.24, 9.27] & 7.06 & +0.52 [-0.07, +0.90] & 96.1 & 71.7 \\
\bottomrule
\end{tabular}
\end{table}



\subsection{Comparison with the Joint Law of \texorpdfstring{\citet{shukor2025scaling}}{Shukor et al.}}

\begin{figure}[!t]
    \centering
    \includegraphics[width=\linewidth]{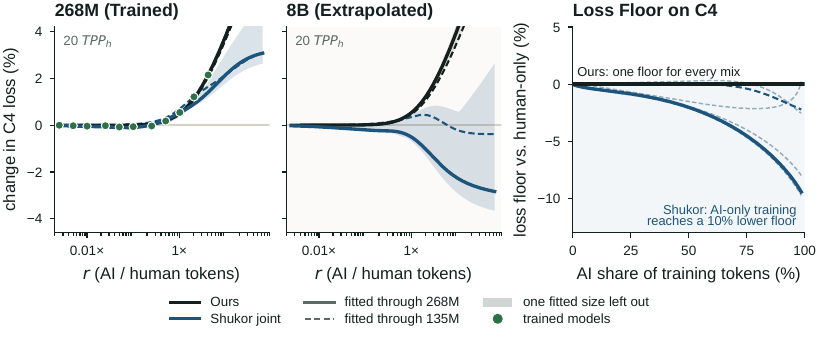}
    \caption{\textbf{The joint law of \citet{shukor2025scaling} extrapolates inconsistently.} Our law and the joint law on C4 at 20 $TPP_h$, fitted through 268M or 135M. Left: 268M, a trained size (points: trained models). Middle: extrapolated to 8B, where the joint law's predicted effect of AI text changes sign with the sizes it is fitted on, while every fit of ours predicts harm. Right: the loss floor each fit of the joint law implies for a training mix, relative to human-only training; our law has one floor for every mix.}
    \label{fig:law-extrapolation}
\end{figure}

The joint law of \citet{shukor2025scaling} is the strongest published scaling law for modeling the effects of AI text. It is second to ours on human text and best on mixed and AI text (\autoref{tab:law-results}, \autoref{tab:law-results-ai-targets}). It has no separate benefit and harm terms, and its fits imply that a model trained on AI text reaches a lower loss floor on human text than one trained on human text (\autoref{fig:law-extrapolation}, right). Its projections beyond the trained sizes are also unstable: fitted through 135M or through 268M, it projects opposite effects of AI text at 8B (\autoref{fig:law-extrapolation}, middle). We trained no 8B model, so these are projections for either law.

\section{Details for the Recommendations}
\label{app:recommendations}

\subsection{Compute-Equivalent Gain (CEG)}
\label{app_sec:ceg}

\begin{figure}[!t]
    \centering
    \includegraphics[width=\linewidth]{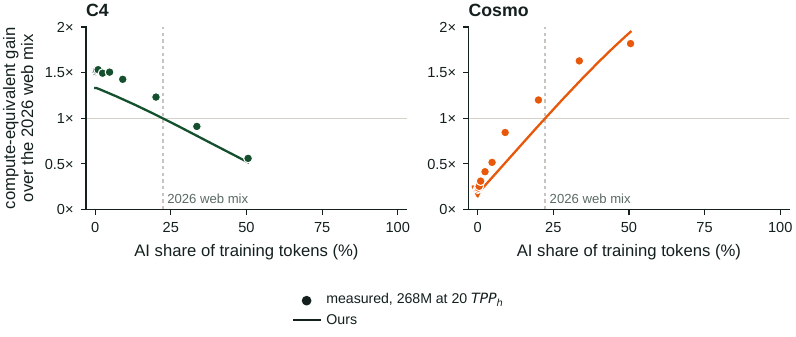}
    \caption{\textbf{Less AI text is always more compute-efficient for C4, and more AI text for Cosmopedia.} Compute-equivalent gain over our 2026 crawl mix (22.3\% AI) of 268M models at 20 $TPP_h$ as the AI share of training tokens varies, on C4 and Cosmopedia.}
    \label{fig:law-ceg-law}
\end{figure}

CEG \citep{davidson2023ceg} provides a direct method to compare the compute efficiency of two models. At a fixed model size, compute is proportional to training tokens ($C\approx6ND$), so a gain in tokens is a gain in compute. If a model trained on $D$ tokens reaches loss $L$, then

\begin{equation}
\mathrm{CEG}_{\mathcal{R}}(D) = \frac{D_{\mathrm{ref}}(L)}{D},
\label{eq:ceg}
\end{equation}

where $D_{\mathrm{ref}}(L)$ is the number of tokens the reference recipe needs to reach the same loss. A gain above one means the recipe is worth more than its compute in reference tokens; below one, less. The law-based version (\autoref{fig:law-ceg-law}) replaces both sides with the fitted law and varies the human budget.

\subsection{Optimal AI Share}
\label{app_sec:optimal-share}
The optimal AI shares in \autoref{fig:law-when-ai} hold the model size $N$ and the human corpus $D_{\mathrm{H}}$ fixed and ask how many AI tokens to add. For each evaluation target we fit the law of \autoref{eq:law} separately and minimize its predicted loss over the ratio of AI to human tokens,
\begin{equation}
r_{\mathrm{opt}}(N, D_{\mathrm{H}}) = \operatorname*{arg\,min}_{0 \le r \le r_{\max}} \widehat{L}\bigl(N, D_{\mathrm{H}}, r\,D_{\mathrm{H}}\bigr),
\qquad
f_{\mathrm{AI}} = \frac{r_{\mathrm{opt}}}{1 + r_{\mathrm{opt}}},
\label{eq:optimal-share}
\end{equation}
where $f_{\mathrm{AI}}$ is the AI share of all training tokens that the figure plots. Only the data term of \autoref{eq:law} depends on $r$, and setting its derivative to zero balances the marginal credit of an AI token against its marginal harm.

\subsection{Repeating Human Tokens vs.\ Adding AI Tokens}
\label{app:repetition}

\begin{figure}[!t]
    \centering
    \includegraphics[width=\linewidth]{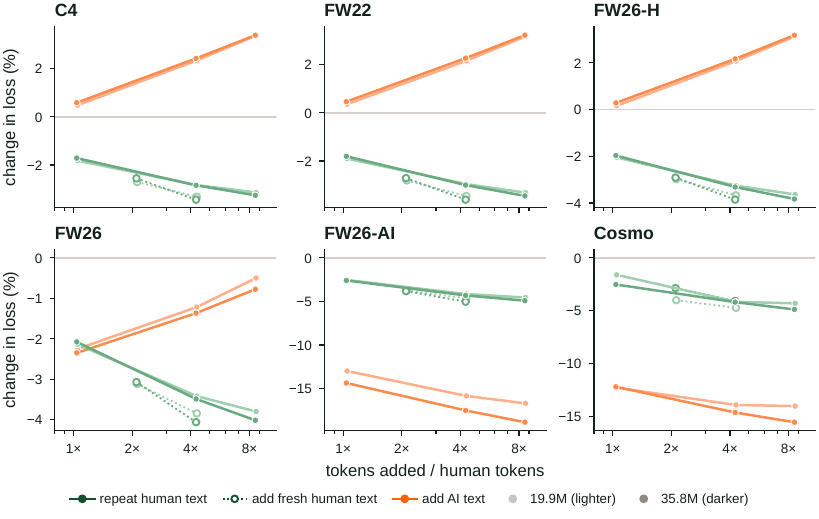}
    \caption{\textbf{Repeating human text beats adding AI text on human-text evaluation sets.} Change in loss on each evaluation set when the human corpus is repeated or the same number of new AI tokens is added, against tokens added per human token. On the human-text sets (C4, FW22, FW26-H) repetition lowers loss where AI text raises it; on the AI-text sets (FW26-AI, Cosmo) AI text lowers loss more.}
    \label{fig:law-repetition-targets}
\end{figure}

Six models (19.9M and 35.8M, 20 $TPP_h$) repeat the human corpus to match $r$ = 1, 4 or 8 and are compared with the six AI additions at the same $r$; the repetition models are not among the 800 of the scaling-law fits.  On the three human-text sets repetition beats adding AI text in all 18 comparisons, lowering loss by 1.7 to 3.8\% while AI text raises it by 0.2 to 3.4\%, and four extra passes recover most of what the same amount of fresh human text gives (see \autoref{fig:law-repetition-targets}). On the two AI-text sets the order reverses: AI text lowers loss by 12 to 19\%, repetition by 1.6 to 4.9\%. On FW26, which mixes the two, AI text wins at $r=1$ and repetition at $r=4$ and 8. 

\subsection{Loss on AI-Generated and Mixed Text}
\label{app:ai-text-targets}

\begin{figure}[!t]
    \centering
    \includegraphics[width=\linewidth]{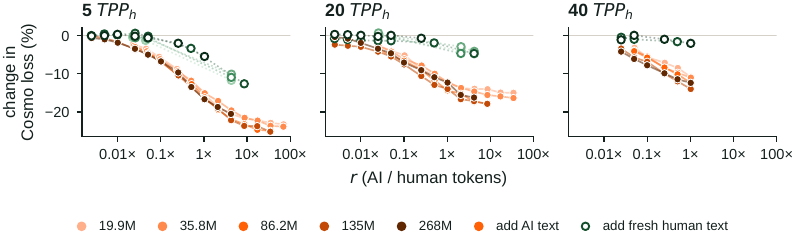}
    \caption{\textbf{On AI-generated text, added AI tokens lower loss at every budget.} Counterpart of \autoref{fig:law-dose-response} for Cosmopedia: change in Cosmo loss when AI tokens or the same number of fresh human tokens are added, at 5, 20 and 40 $TPP_h$ for every fitted size. AI text lowers the loss at every budget and size, by up to 25\%; fresh human text lowers it far less.}
    \label{fig:law-dose-response-cosmopedia}
\end{figure}

\begin{figure}[!t]
    \centering
    \includegraphics[width=0.72\linewidth]{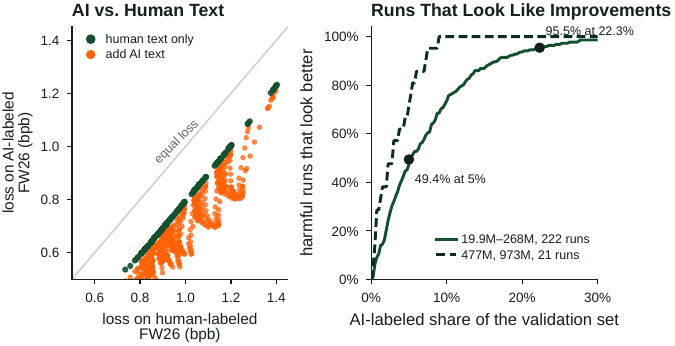}
    \caption{\textbf{Mixed validation data hides harm to human text.} Left: every model's loss on the AI-labeled against the human-labeled partition of FW26. Human-only models have 12 to 27\% lower loss on AI text, and models trained with AI text fall further below the equal-loss line. Right: of the 243 AI additions that raise loss on human-labeled text, the share a validation set reports as improvements as the AI-labeled share of its text grows, for the fitted sizes and the held-out sizes: 49.4\% at 5\% AI and 95.5\% at our 2026 crawl's 22.3\%.}
    \label{fig:ai-text-eval}
\end{figure}

\autoref{fig:law-dose-response-cosmopedia} shows how our models learn on AI text. \autoref{fig:ai-text-eval} shows that models have lower loss on AI-labeled text than on human-labeled text, more so when trained with AI text, and that a mixed validation set can hide runs that look like improvements but raise loss on human text.

\section{Downstream Metrics}
\label{app:downstream}
\label{app:beyond-loss}

\begin{figure}[!t]
    \centering
    \includegraphics[width=0.62\linewidth]{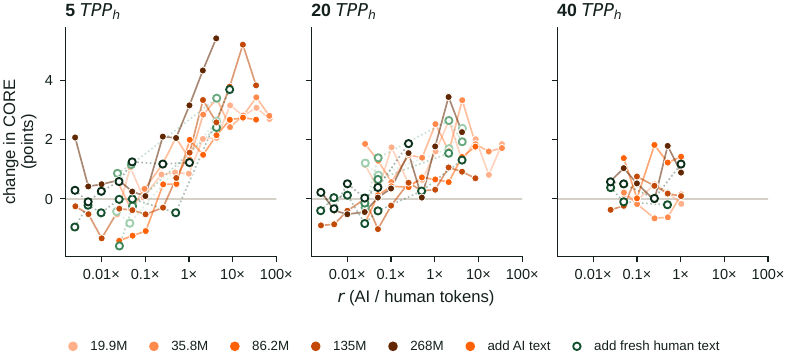}
    
    \caption{\textbf{Added AI tokens raise CORE about as much as fresh human tokens.} Change in CORE  against each model's human-only control, as AI tokens or the same number of fresh human tokens are added. MMLU \citep{hendrycks2021measuring} is not drawn because it stays at chance accuracy for every model, as expected at these sizes.}
    \label{fig:downstream-core-mmlu}
\end{figure}

We find that CORE \citep{li2024dclm} performance increases similarly when models are trained on additional AI or human tokens, as depicted in \autoref{fig:downstream-core-mmlu}. We hypothesize that while AI text may be good at these standard tasks, there will be greater differences in more
unique tasks that may require diversity or creativity,
where AI-written text is known to diverge from
human writing \citep{russell2026storyscope}.

\paragraph{Same benchmark scores, different fit to human text.} Benchmark accuracy and fit to human text come apart. At 20 $TPP_h$, adding at least as many tokens as the human corpus holds raises CORE by 1.6 points with AI text and by 1.9 points with fresh human text (23 and 6 runs), yet C4 loss moves in opposite directions: up 2.3\% with AI text and down 3.0\% with fresh human text. Models trained with AI text answer benchmark questions about as well, but they model human text less closely and write more like AI (\autoref{tab:generation-behavior}). CORE's tasks reward knowledge that AI web text also carries, while loss on human text also measures how well a model matches the formats and style of human writing, where AI text differs most (\autoref{fig:web-formats-filters}, left).

\begin{figure}[!t]
    \centering
    \includegraphics[width=0.85\linewidth]{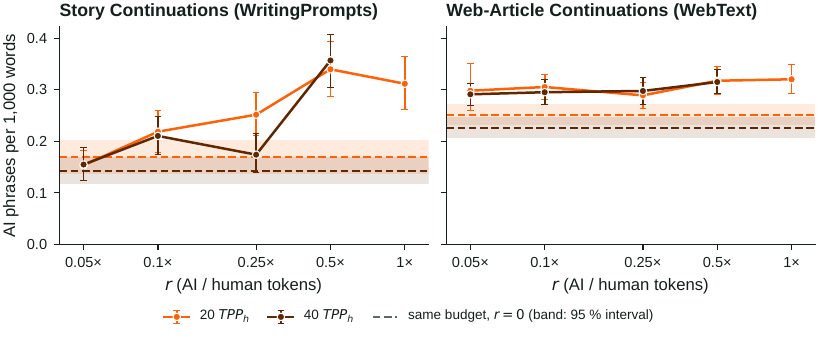}
    \caption{
    \textbf{Models trained with AI text use more AI-typical phrases.} Rate of 26 AI-typical phrases per 1,000 words in generations from models trained with AI text added at ratio $r$.}
    \label{fig:ai-phrase}
\end{figure}

\paragraph{AI-typical phrases increase.}\label{app:qualitative} We generate continuations with the twelve held-out 973M models (three human-only controls and nine AI additions) for 500 WritingPrompts \citep{fan2018hierarchical} story prompts and for WebText \citep{radford2019language} article openings, and count 26 phrases typical of AI-generated text (such as ``delve'', ``tapestry'' and ``it is important to note''). In story continuations the rate roughly doubles by $r=0.5$, from 0.14--0.17 to 0.34--0.36 per 1,000 words at both budgets (\autoref{fig:ai-phrase}). The 268M ladders extend $r$ to 4 (\autoref{tab:generation-behavior}). At 20 $TPP_h$ the phrase rate keeps rising, from 0.16 per 1,000 words without AI text to 0.36 at $r=1$ and 0.54 at $r=4$, and at 5 $TPP_h$ it reaches 0.72 at $r=4$. Pangram~3.3.2 labels 18.6\% of the human-only model's stories AI, since base models often drift into generic web prose, and 28.4\% at $r=1$ and 37.2\% at $r=2$, twice the control; at $r=4$ the share falls back to 28.9\%.

\begin{table}[t]
\centering
\small
\setlength{\tabcolsep}{6pt}
\caption{\textbf{The 268M models trained with more AI text write more like AI.} Story continuations of 500 WritingPrompts prompts from the 268M models at 5 and 20 $TPP_h$, one model per $r$. AI-typical phrases per 1,000 words over eight continuations per prompt, and the share of one continuation per prompt that Pangram~3.3.2 labels AI, with 95\% intervals from a bootstrap over prompts.}
\label{tab:generation-behavior}
\begin{tabular}{rrccc}
\toprule
 & & \multicolumn{2}{c}{AI phrases per 1,000 words} & Stories labeled AI (\%) \\
\cmidrule(lr){3-4}\cmidrule(lr){5-5}
$r$ & AI share (\%) & 5 $TPP_h$ & 20 $TPP_h$ & 20 $TPP_h$ \\
\midrule
0 & 0.0 & 0.16 & 0.16 & 18.6 [15.4, 22.0] \\
0.025 & 2.4 & 0.24 & 0.18 & 22.4 [18.6, 26.1] \\
0.05 & 4.8 & 0.18 & 0.21 & 21.1 [17.7, 24.7] \\
0.25 & 20.0 & 0.26 & 0.24 & 26.8 [23.0, 30.6] \\
0.5 & 33.3 & 0.23 & 0.24 & 22.7 [19.1, 26.7] \\
1 & 50.0 & 0.28 & 0.36 & 28.4 [24.4, 32.4] \\
2 & 66.7 & 0.41 & 0.46 & 37.2 [33.0, 41.6] \\
4 & 80.0 & 0.72 & 0.54 & 28.9 [25.1, 33.1] \\
\bottomrule
\end{tabular}
\end{table}

\end{document}